\documentclass[letterpaper, 10 pt, conference]{ieeeconf}  

\IEEEoverridecommandlockouts                        
\title{\LARGE \bf
    $\mathcal{T(R, O)}$ Grasp: Efficient Graph Diffusion of Robot-Object Spatial Transformation for Cross-Embodiment Dexterous Grasping
}

\author{
  Xin Fei\textsuperscript{1,2*},
  Zhixuan Xu\textsuperscript{1,2*},
  Huaicong Fang\textsuperscript{3},
  Tianrui Zhang\textsuperscript{1},
  Lin Shao\textsuperscript{1,2\textdagger}\\
  \textsuperscript{1}Department of Computer Science, National University of Singapore\quad \textsuperscript{2}RoboScience\\
  \textsuperscript{3}College of Control Science and Engineering, Zhejiang University\\
  \thanks{\textsuperscript{*} denotes equal contribution}
  \thanks{\textsuperscript{\textdagger} denotes corresponding author}
  \vspace{-40pt}
}

\usepackage{bbm}
\usepackage{etoolbox}
\usepackage{multicol}
\usepackage{cite}
\usepackage[bookmarks=true]{hyperref}
\usepackage{graphics} % for pdf, bitmapped graphics files
\usepackage{times} % assumes a new font selection scheme installed
\usepackage{amsmath} % assumes amsmath package installed
\usepackage{amssymb}  % assumes amsmath package installed
\usepackage{svg}
\usepackage{mathrsfs}
\usepackage{amsfonts}
\usepackage{wrapfig}
\usepackage{amsmath}

\usepackage{kantlipsum}
\usepackage{subcaption}
\usepackage{marvosym}

\usepackage{graphicx}
\usepackage{float}
\usepackage{ctable}
\usepackage{cuted}
\usepackage{colortbl}
\usepackage{multirow}
\usepackage{wrapfig}
\usepackage[misc,geometry]{ifsym}
\usepackage{algorithm}
\usepackage{algpseudocode}
\usepackage{xcolor}   % For color
\usepackage{pifont}    % For check and cross symbols

\usepackage{bm}
\usepackage{stfloats}
\usepackage{caption}
\usepackage{setspace}
\let\labelindent\relax
\usepackage{enumitem}
\usepackage{threeparttable}
\newcommand{\TODO}[1][]{\textcolor{red}{\bf [TODO]}}
\setlist[enumerate,1]{itemsep=3pt}

\definecolor{formalgreen}{rgb}{0.1, 0.7, 0.1}  % A darker, more formal green
\definecolor{formalred}{rgb}{0.9, 0.2, 0.2}  % A darker, more formal green

\begin{document}

\maketitle
\thispagestyle{empty}
\pagestyle{empty}

\begin{strip}
    \centering
    \includegraphics[width=\linewidth]{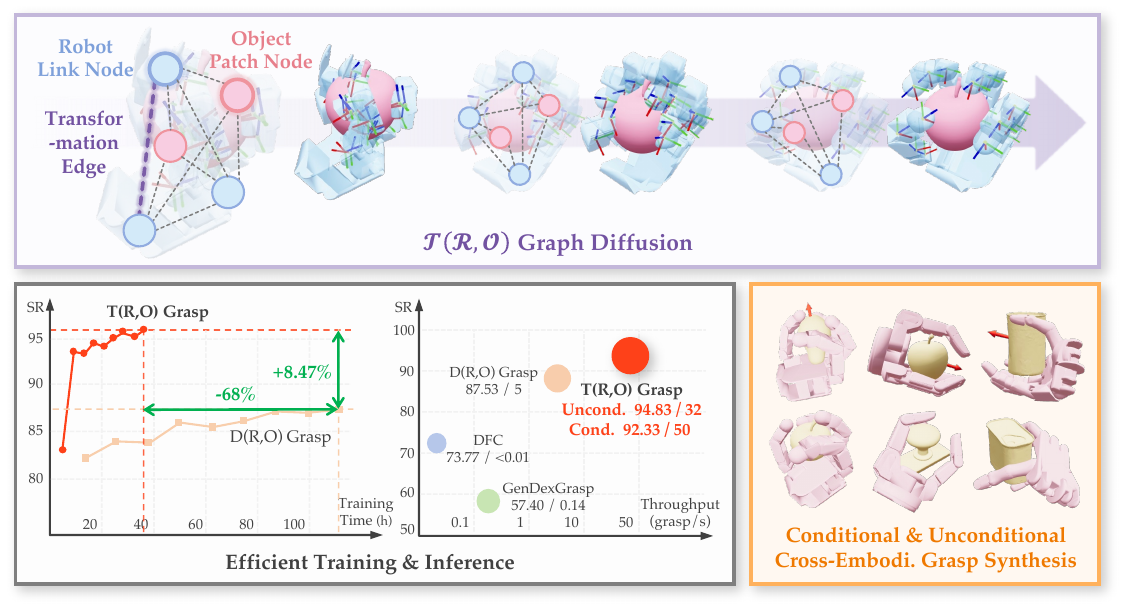}
    \captionof{figure}{Given object point cloud and hand URDF, $\mathcal{T(R,O)}$ Grasp efficiently supports both conditioned and unconditioned grasp synthesis utilizing a graph diffusion model. Compared with $\mathcal{D(R,O)}$ Grasp~\cite{wei2024dro}, our method achieves superior performance with lower memory usage, significantly higher inference speed and throughput.}
    \vspace{5pt}
    \label{fig:teaser}
\end{strip}

%%%%%%%%%%%%%%%%%%%%%%%%%%%%%%%%%%%%%%%%%%%%%%%%%%%%%%%%%%%%%%%%%%%%%%%%%%%%%%%%
\begin{abstract}
Dexterous grasping remains a central challenge in robotics due to the complexity of its high-dimensional state and action space. We introduce $\mathcal{T(R,O)}$ Grasp, a diffusion-based framework that efficiently generates accurate and diverse grasps across multiple robotic hands. At its core is the $\mathcal{T(R,O)}$ Graph, a unified representation that models spatial transformations between robotic hands and objects while encoding their geometric properties. A graph diffusion model, coupled with an efficient inverse kinematics solver, supports both unconditioned and conditioned grasp synthesis. Extensive experiments on a diverse set of dexterous hands show that $\mathcal{T(R,O)}$ Grasp achieves average success rate of 94.83\%, inference speed of 0.21s, and throughput of 41 grasps per second on an NVIDIA A100 40GB GPU, substantially outperforming existing baselines. In addition, our approach is robust and generalizable across embodiments while significantly reducing memory consumption. More importantly, the high inference speed enables closed-loop dexterous manipulation, underscoring the potential of $\mathcal{T(R,O)}$ Grasp to scale into a foundation model for dexterous grasping. The code, appendix, and videos are available at~\href{https://nus-lins-lab.github.io/trograspweb/}{https://nus-lins-lab.github.io/trograspweb/}
.
\end{abstract}
%%%%%%%%%%%%%%%%%%%%%%%%%%%%%%%%%%%%%%%%%%%%%%%%%%%%%%%%%%%%%%%%%%%%%%%%%%%%%%%%
\section{Introduction}
\label{sec.1}
Grasping with dexterous hands is a fundamental capability for achieving precise, human-level manipulation. Yet, efficiently generating diverse and high-quality grasps remains a longstanding challenge, largely due to the high dimensionality of dexterous hands and the difficulty of ensuring both stability and precision. Existing learning-based research has primarily explored two types of representations: \emph{robot-centric} and \emph{object-centric}. \emph{Robot-centric} approaches~\cite{wan2023unidexgrasp++, dfc, jiang2021graspTTA, huang2025fungrasp, liu2024realdex}, such as representing grasp with wrist poses or joint values, enable rapid inference by mapping observations directly to control signals. However, they often suffer from poor sample efficiency and limited transferability, as the learned representations are implicitly tied to specific hand embodiments and fail to generalize across novel morphologies. In contrast, \emph{object-centric} paradigms focus on describing grasps in terms of object features such as contact points or affordances~\cite{shao2020unigrasp, xu2024manifoundation, xu2023unidexgrasp, li2023gendexgrasp, varley2015generating, attarian2023geometry, fang2025anydexgrasp, zhao2024graingrasp}. This representation naturally promotes transferability across diverse objects and robotic platforms. However, it is often brittle under partial observations and typically requires an additional optimization stage, such as fingertip IK or constrained fitting of contact maps, to convert abstract predictions into feasible joint configurations, making the process computationally expensive and time-consuming. 

Recently, $\mathcal{D(R,O)}$ Grasp~\cite{wei2024dro} introduced an \emph{interaction-centric} representation for dexterous grasping. Unlike \emph{robot-centric} or \emph{object-centric approaches}, an \emph{interaction-centric representation} characterizes a grasp by directly modeling the spatial relationships between the robot hand and the object, embedding both entities into a shared interaction space. This formulation combines the strengths of robot-centric and object-centric paradigms: it retains embodiment awareness while capturing transferable object-level features. As a result, it facilitates generalization across different hand embodiments and object shapes, preserves the fine-grained geometric structure of potential contacts, and remains more robust under partial observations. In $\mathcal{D(R,O)}$, this is achieved by constructing a distance matrix between points sampled on the hand and the object. However, the point-to-point feature maps in $\mathcal{D(R,O)}$ Grasp require substantial GPU memory. Thus, the representation is resource-intensive and difficult to scale up. In addition, $\mathcal{D(R,O)}$ relies on a suitable initial status as input, which makes the model performance vulnerable to infeasible initialization.

To overcome the limitations of $\mathcal{D(R,O)}$ representation, we propose $\mathcal{T(R,O)}$ Grasp, an efficient graph diffusion model for cross-embodiment dexterous grasp synthesis based on $\mathcal{T(R,O)}$ Graph representation. Formally, for any object-hand pair, we formulate a unified \textbf{T}ransformation Graph between the \textbf{R}obot and \textbf{O}bject, namely the $\mathcal{T(R,O)}$ Graph. In $\mathcal{T(R,O)}$ graph, nodes encode object and link geometries together with their spatial transformations in object global frame for a grasp, while edges consist of relative spatial transformations between nodes. Compared to \emph{robot-centric} representation, such definition is unified and generalizable across all objects and dexterous hand embodiments. Unlike $\mathcal{D(R,O)}$ that requires memory-intensive point-to-point representation, $\mathcal{T(R,O)}$ Graph establishes relations between object patches and hand links, which significantly reduces memory usage and enables faster training and inference.

Built upon $\mathcal{T(R,O)}$ representation, we implement an efficient transformer-based graph diffusion model for $\mathcal{T(R,O)}$ Graph generation, thereby enabling cross-embodiment dexterous grasp synthesis either unconditionally or conditioned on a hand status. In extensive experiments, $\mathcal{T(R,O)}$ Grasp achieves the grasping success rate of $94.83\%$, with an average inference speed of $0.21$s and a throughput of $41$ grasps per second on a NVIDIA A100 40GB GPU. More importantly, the fast inference speed allows $\mathcal{T(R,O)}$ Grasp to support \emph{closed-loop dexterous grasping and manipulation}, a key requirement for real-world deployment. In conclusion, our main contributions are summarized as:

\begin{enumerate}
    \item We introduce the $\mathcal{T(R,O)}$ Graph, a unified and generalizable representation for modeling interactions across all object–hand pairs. This representation is significantly more memory-efficient than the $\mathcal{D(R,O)}$ formulation and establishes a foundation for scaling toward large dexterous grasping foundation models.
    \item We develop an efficient graph-diffusion model built on the $\mathcal{T(R,O)}$ Graph that supports both conditioned and unconditioned grasp synthesis. This design not only mitigates sensitivity to infeasible initializations but also enables reliable closed-loop dexterous manipulation.
    \item We conduct extensive experiments both in simulation and real world, demonstrating that $\mathcal{T(R,O)}$ Grasp significantly outperforms all baselines on success rate for dexterous grasping while delivering substantially higher efficiency than existing methods.
\end{enumerate}
%%%%%%%%%%%%%%%%%%%%%%%%%%%%%%%%%%%%%%%%%%%%%%%%%%%%%%%%%%%%%%%%%%%%%%%%%%%%%%%%
\section{Related Work}
\subsection{Representation for Dexterous Grasping}
\label{sec:2.1}

Grasping with dexterous robotic hands holds great potential for achieving human-like precision in manipulation. However, it remains highly challenging due to the inherent high dimensions. Achieving a unified model for cross-embodiment dexterous grasping requires an efficient and robust representation that consistently encodes the geometry and interactions of objects and robotic hands. A natural and straightforward approach is to represent a grasp with hand joint values~\cite{wan2023unidexgrasp++, dfc, jiang2021graspTTA, 9710672}. While these predictions can be directly executed, the representation of joint values lacks a canonical definition, as a single hand can admit multiple valid joint parameterizations. In addition, its generalization across embodiments is limited, and its efficiency is hindered by the prohibitive computational cost of reinforcement learning in high-dimensional action space.

Another stream of research formulates grasp representation in an object-centric way, typically by encoding object contact points or contact maps~\cite{shao2020unigrasp, xu2024manifoundation, xu2023unidexgrasp, li2023gendexgrasp, varley2015generating, attarian2023geometry, liu2023contactgen}. Though these methods allows generalization across embodiments and objects, they typically rely on contact-based optimization, which hinders real-time grasp synthesis. More recently, $\mathcal{D(R,O)}$ Grasp~\cite{wei2024dro} pioneers to introduce an interaction-based representation by predicting a point-to-point distance matrix between objects and robotic hands, which significantly facilitates robust generalization across diverse objects and hand embodiments. However, the $\mathcal{D(R,O)}$ representation faces several limitations: (1) the point-to-point feature maps constructed prior to $\mathcal{D(R,O)}$ estimation incur substantial memory overhead, which is memory-inefficient and hinders its scalability; (2) the prediction relies heavily on the initial pose (as it conditions on latent outputs from a CVAE~\cite{sohn2015learning}), which can degrade performance when the initial pose is suboptimal or infeasible for a valid grasp; (3) $\mathcal{D(R,O)}$ representation requires additional optimization to obtain $\text{SE(3)}$ transformations of each link. To overcome these limitations, we propose $\mathcal{T(R,O)}$ Graph as a unified and efficient representation which centers on spatial transformation between hand links and object patches. In addition, we implement a lightweight graph diffusion model to support both unconditioned and conditioned grasp synthesis.

\subsection{Grasp Synthesis with Generative Models}
\label{sec:2.2} With the success of generative models, numerous researches train generators on large-scale datasets with sample refinement to perform grasp synthesis. Contact-GraspNet~\cite{sundermeyer2021contact} efficiently generates collision-free grasps in cluttered scenes, while GraspME~\cite{9515479} introduces a manifold estimator to model feasible grasp in object manifold. Neural Grasp Distance Fields~\cite{10160217} leverages implicit distance representations for continuous grasp synthesis, and SE(3)-DiffusionFields~\cite{urain2022se3dif} further employs diffusion-based learning to jointly optimize grasp and motion through smooth cost functions. Despite these advances, all prior methods remain constrained to 6-DoF pose of a single robotic gripper, which is fundamentally inadequate for dexterous hand grasping due to the lack of a suitable representation. In this paper, our proposed $\mathcal{T(R,O)}$ Graph provides a unified and lightweighted representation for dexterous hands with arbitrary geometry and numbers of links, enabling high-dimensional grasp synthesis with generative models far beyond the 6-DoF robotic gripper.
%%%%%%%%%%%%%%%%%%%%%%%%%%%%%%%%%%%%%%%%%%%%%%%%%%%%%%%%%%%%%%%%%%%%%%%%%%%%%%%%
\section{Method\label{sec:method}}
\noindent Given an object point cloud $P^O \in \mathbb{R}^{N \times 3}$, URDF description of the dexterous hand, and the point clouds of $L$ links $\{P_i^R\}_{i=1}^L$ in the local frame of each link, $\mathcal{T(R,O)}$ Grasp generates diverse grasps by first predicting $\text{SE(3)}$ transformations of all links while temporarily relaxing joint constraints. Then, the predicted transformations are refined and converted into executable joint values through Pyroki-based inverse kinematics~\cite{pyroki2025}, where joint constraints are enforced to ensure feasibility.

\begin{figure*}[t] \centering
    \includegraphics[width=\linewidth]{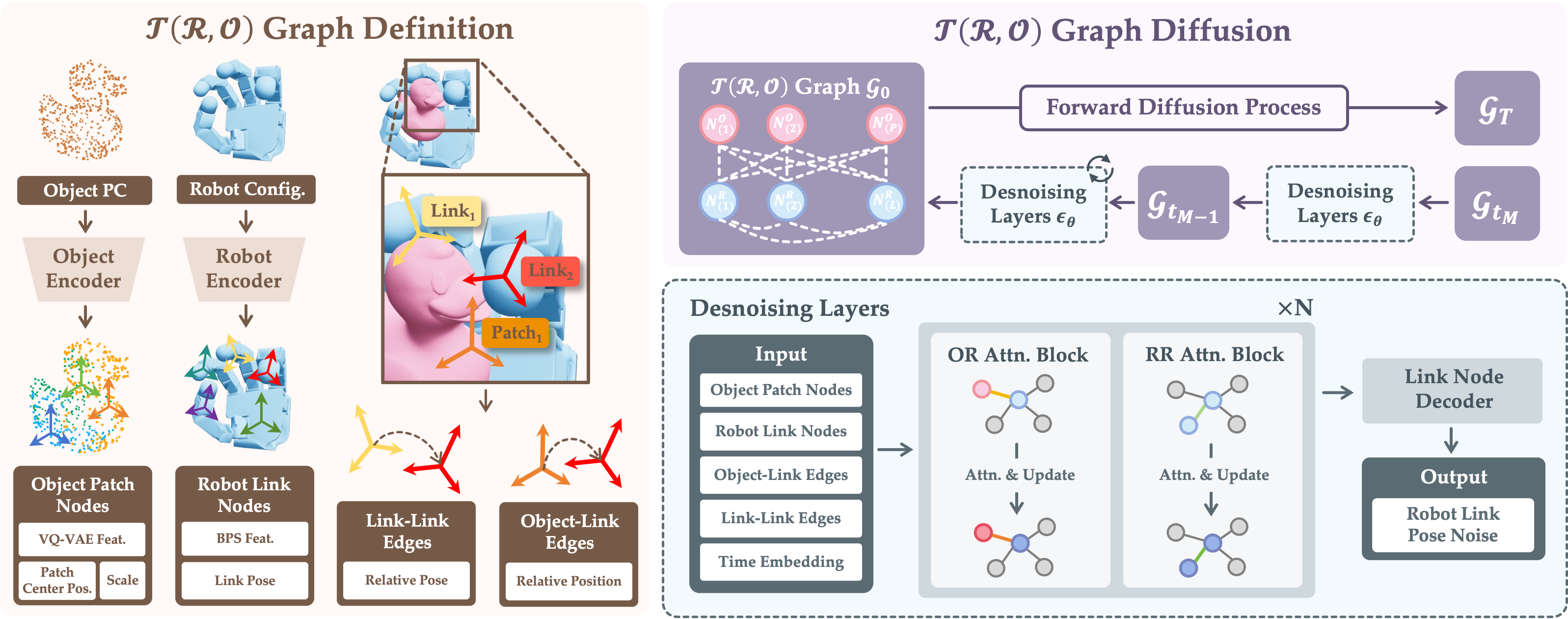}
    \caption{Overview of $\mathcal{T(R,O)}$ Grasp: We first define $\mathcal{T(R,O)}$ Graph to represent spatial transformations of robotic links and objects with auxiliary geometry information. Next, we introduce a graph diffusion model that enables both unconditioned and conditioned grasp synthesis.}
    \label{fig:pipeline}
    \vspace{-15pt}
\end{figure*}

\noindent \textbf{Method Overview.} An overview of our proposed framework is illustrated in Fig.~\ref{fig:pipeline}. We formulate dexterous grasp synthesis as the denoising process of our proposed $\mathcal{T}(R,O)$ Graph (Sec.~\ref{sec:3.1}). Since spatial transformations are defined consistently across different objects and hand embodiments, this graph serves as a unified representation of a diffusion-based generative process (Sec.~\ref{sec:3.2}), where noisy link poses are progressively denoised into valid grasps. Furthermore, we elaborate on the training and inference process of $\mathcal{T(R,O)}$ Grasp in Sec.~\ref{sec:3.3}, which enables efficient and accurate grasp synthesis both unconditionally (without external constraints) and conditionally (guided by an initial hand status). 

\subsection{$\mathcal{T(R,O)}$ Graph}
\label{sec:3.1}
Learning cross-embodiment dexterous grasping requires a unified representation to characterize the relation between dexterous hands and objects. Compared to joint values, which vary substantially across embodiments, spatial transformations offer a more robust and consistent representation cross different objects and embodiments. Therefore, we construct $\mathcal{T(R,O)}$ Graph based upon spatial $\text{SE(3)}$ transformation of object patches and dexterous hand links. In $\mathcal{T(R,O)}$ Graph, nodes represent the geometric property of each object patch or hand link, along with its spatial transformation with respect to the object global frame for a grasp. Edges encode the relative transformations between object patches and hand links. 

Since $\text{SE(3)}$ is a non-linear manifold, direct application of the standard diffusion process is not feasible. Therefore, we adopt the Lie group representation of $\text{SE(3)}$ to model spatial transformations throughout our framework. According to~\cite{sola2018micro}, the Lie algebra of $\text{SE(3)}$ is isomorphic to the vector space $\mathbb{R}^6$, enabling transformation between the $\text{SE(3)}$ Lie group and vector space $\mathbb{R}^6$ via exponential and logarithmic mapping operations as follows:

\vspace{-3mm}
\begin{equation}
\label{eq:0}
    \mathbf{T} = \exp\!\left(\boldsymbol{\psi}\right), \ 
    \boldsymbol{\psi} = \log\!\left(\mathbf{T}\right),
\end{equation}
\vspace{-5mm}

where $\boldsymbol{\psi} \in \mathbb{R}^6, \ \mathbf{T} \in \text{SE(3)}$ denote rotation vector and homogeneous transformation matrix, respectively. Hence, we represent $\text{SE(3)}$ transformations as 6D rotation vectors in our $\mathcal{T(R,O)}$ Graph during the diffusion process, and map them back to $\text{SE(3)}$ after inference. In the following, we describe the detailed components of the $\mathcal{T(R,O)}$ Graph.

\noindent \textbf{Object Nodes.} Given an object point cloud, we first partition it into $P$ patches using farthest point sampling. To capture the spatial and geometric characteristics of the object to be grasped, we employ the pretrained VQ-VAE~\cite{van2017neural} encoder for 3D fracture assembly in~\cite{wang2025puzzlefusionpp} to extract geometry tokens $\{f_i^O\}_{i=1}^P$ from the normalized patches within a unit sphere. Object nodes are then constructed by concatenating patch center locations $\{c_i^O\}_{i=1}^P$, global object scale $s^O$, and corresponding patch geometry tokens as Eq.~\ref{eq:1}, where $d^O$ denotes feature dimension of object nodes.

\vspace{-5mm}
\begin{equation}
N^O = \{\text{concat}(c_i^O, s^O, f_i^O)|i=1,\dots,P\} \in \mathbb{R}^{P \times d^O},
\label{eq:1}
\end{equation}
\vspace{-5mm}

\noindent \textbf{Link Nodes.} Link nodes encode both the geometric properties of each link and its spatial transformation with respect to the object’s global frame for each grasp. To represent geometry, we apply the Basis Point Set~(BPS) algorithm~\cite{prokudin2019efficient} to the link point clouds $\{P_i^R\}_{i=1}^L$. Specifically, $B$ basis points are uniformly sampled within a unit sphere, and each link point cloud is normalized into the same unit sphere. The BPS feature of a link is then computed as a vector of length $B$, where each entry corresponds to the minimum distance between the point cloud and a basis point. This encoding produces fixed-length features invariant to the number of input points, and can be implemented efficiently in PyTorch, making it both scalable and generalizable across diverse dexterous hand embodiments.

Next, an MLP-based encoder embeds the BPS feature together with the link center $\{c_i^R\}$ and scale $\{s_i^R\}$ to obtain the geometry embedding for each link. The geometry embedding is further concatenated with the 6D rotation vector $\{\psi_i^R\}$ of the link global pose with respect to the object frame in each state to formulate the link nodes within the graph:

\vspace{-5mm}
\begin{align}
b_i &= \text{MLP}(\text{BPS}(P_i^R), c_i^R, s_i^R), \ i=1,\dots,L, \\
N^R &= \{\text{concat}(\psi_i^R, b_i)|i=1,\dots,L\} \in \mathbb{R}^{L \times d^R},
\label{eq:2}
\end{align}
\vspace{-5mm}

\noindent \textbf{Edges.} Since the relative poses of object nodes are fixed for rigid body grasping, no edges are defined between object nodes. Instead, we define edges only between object nodes and link nodes, as well as between link nodes. Notably, we establish complete connection for stronger spatial transformation constraints. Link-link edges are defined as the relevant spatial transformation between two link nodes. To ensure consistency across different objects, we assign all object transformation matrices $\{T^O_i\}_{i=1}^P$ an identity rotation $R^O = I$ and only use its patch center $\{c^O_i\}_{i=1}^P$ as translation. The construction of graph edges is then formulated as:

\vspace{-3mm}
\begin{align}
\label{eq:3}
    T^{OR}_{i,j} &= (T^O_i)^{-1} T^R_j, &
    T^{RR}_{j,k} &= (T^R_j)^{-1} T^R_k, \\
    E^{OR} &= \{\log(T^{OR}_{i,j})\}, &  
    E^{RR} &= \{\log(T^{RR}_{j,k})\},
\end{align}
\vspace{-3mm}

where $E^{OR} \in \mathbb{R}^{6 \times P \times L}$ and $E^{RR} \in \mathbb{R}^{6 \times L(L-1)/2}$ represent object-link edges and link-link edges, respectively.

Integrating spatial transformations between object patches and links with rich geometry embeddings, our proposed $\mathcal{T(R,O)}$ Graph is defined in Eq.~\ref{eq:4}, which serves as the foundation structure for subsequent graph diffusion.

\vspace{-3mm}
\begin{equation}
\mathcal{G}=(N^O, N^R, E^{OR}, E^{RR}).
\label{eq:4}
\end{equation}

\subsection{$\mathcal{T(R,O)}$ Graph Diffusion Model}
\label{sec:3.2}
We employ a graph diffusion model that operates on our proposed $\mathcal{T(R,O)}$ Graph, enabling the generation of diverse and precise grasps from random noise that can generalize across different objects and embodiments. Diffusion Probabilistic Models (DDPM)~\cite{ho2020denoising} generate data by gradually denoising Gaussian noise through a stochastic process, but require many inference steps. To achieve more efficient inference, we adopt the Denoising Diffusion Implicit Model (DDIM)~\cite{song2020denoising}, a variant of DDPM that accelerates sampling, allows controllable generation diversity, and naturally supports conditioned generation.

\vspace{1mm}
\noindent \textbf{Forward Process.} As illustrated in Sec.~\ref{sec:3.1}, object nodes ($N^O$) remain constant during both training and inference, whereas edges ($E^{OR}, E^{RR}$) have to satisfy the spatial constraints between nodes. Consequently, noise injection and denoising are applied exclusively to link poses $\Psi^R = \{\psi_i^R\}$. Given a noise-free graph $\mathcal{G}_0=(N^O, N_0^R, E_0^{OR}, E_0^{RR})$ derived from a valid grasp, we first gradually add Gaussian noise with variance schedule $\beta_1 < \dots < \beta_T$ to the link poses as follows:

\vspace{-2mm}
\begin{equation}
p(\Psi_t^R|\Psi_{t-1}^R) := \mathcal{N}(\sqrt{1-\beta_t} \Psi_{t-1}^R, \beta_tI),
\label{eq:5}
\end{equation}
\vspace{-5mm}

Similar to~\cite{ho2020denoising}, link poses at arbitrary timestamp $t$ can be directly sampled from the noisy-free graph (as illustrated in Eq.~\ref{eq:5}), where $\alpha_t := 1 - \beta_t$ and $\bar{\alpha}_t := \prod_{s=1}^t \alpha_s$.

\vspace{-3mm}
\begin{equation}
p(\Psi_t^R|\Psi_0^R) := \mathcal{N}(\sqrt{\bar{\alpha}_t} \Psi_0^R, (1-\bar{\alpha}_t)I),
\label{eq:6}
\end{equation}
\vspace{-5mm}

The noisy link nodes $N_t^R$ are obtained by concatenating the noisy pose vectors with the original BPS embeddings. With the noisy link nodes at timestamp $t$, the corresponding edges are recomputed as the relative spatial transformation between graph nodes:

\vspace{-5mm}
\begin{align}
N_t^R &= \{\text{concat}((\psi^R_{i, t}, b_i)|i=1,\dots,L)\}, \\
E_t^{OR} &= \text{Edge}(N^O, N_t^R), \ E_t^{RR} = \text{Edge}(N_t^R),
\label{eq:7}
\end{align}
\vspace{-5mm}

where $\text{Edge}(\cdot)$ denotes the operation that generates edges from the given graph nodes. Thus, the noisy $\mathcal{T}(R,O)$ Graph at step $t$, $\mathcal{G}_t = (N^O, N_t^R, E_t^{OR}, E_t^{RR})$, is derived from the noisy-free graph $\mathcal{G}_0$.

\vspace{1mm}
\noindent \textbf{Denoising Model.} To predict the link node noise from a noisy $\mathcal{T(R,O)}$ Graph $\mathcal{G}_t$, we implement a transformer-based graph denoising model $\epsilon_{\theta}(\mathcal{G}_t, t)$. Each graph element is first encoded into token representation 
$\{z^O, z^R, z^{OR}, z^{RR}\}$ for object nodes, link nodes, object-link and link-link edges, 
augmented with timestamp embedding $\phi(t)$. These tokens are then processed by a stack of graph denoising layers, where Object–Robot (OR) and Robot–Robot (RR) attention mechanisms iteratively update link nodes and all edges.

In both OR and RR attention, target nodes ($X_{\text{tgt}}$) and connecting edges ($E$) are refined by aggregating information from the corresponding source nodes ($X_{\text{src}}$). For OR attention, we set $X_{\text{src}}=z^O, X_{\text{tgt}}=z^R$ and $E=E^{OR}$, incorporating object geometry and spatial information into link updates. While for RR attention, we set $X_{\text{src}}=X_{\text{tgt}}=z^R$ with $E=E^{RR}$, allowing the model to reason link-wise spatial relations. Formally, the attention process can be formulated as Eq.~\ref{eq:8}, where $h_Q, h_K, h_V$ are learnable projections, and $\text{attn}(\cdot)$ denotes attention-based aggregation. After all denoising layers update the graph, the output head aggregates the intermediate graph representations from all layers to predict the noise on link nodes, which is then used to recover clean link poses for valid grasps. A detailed figure of attention structure is provided in the Appendix.

\vspace{-15pt}
\begin{align}
Q &= h_Q(X_{\text{tgt}}) \in \mathbb{R}^{N_{\text{tgt}} \times d}, \nonumber \\
K &= h_K(X_{\text{src}}) \in \mathbb{R}^{N_{\text{src}} \times d}, \nonumber \\
V &= h_V(\text{concat}(X_{\text{tgt}}, X_{\text{src}}, E)) \in \mathbb{R}^{N_{\text{tgt}} \times N_{\text{src}} \times d}, \label{eq:8}\\
X_{\text{tgt}} &\leftarrow \text{MLP}(X_{\text{tgt}} + \text{attn}(Q, K, V)), \nonumber \\
E &\leftarrow \text{MLP}(V, E), \nonumber
\end{align}
\vspace{-15pt}

\noindent \textbf{Reverse Process.} Following the DDIM inference strategy, $\mathcal{T(R, O)}$ Graph Diffusion Model learns transitions $p_{\theta}(G_{t_{i-1}}|G_{t_i})$ from random noisy $\mathcal{T(R,O)}$ Graph, where $\theta$ denotes the learnable parameters, and $\mathcal{T}^S=\{t_i\}_{i=1}^M$ is an ascending sequence of diffusion timesteps. According to~\cite{song2020denoising}, the distribution of link poses $(\Psi^R)$ in the reverse process can be formulated as:

\vspace{-4mm}
\begin{align}
\Psi^R_{t'} &= 
\sqrt{\bar{\alpha}_{t'}}\hat{\Psi}^R_0
+ \sqrt{1-\bar{\alpha}_{t'}- \sigma_t^2}
  \epsilon_{\theta}(\mathcal{G}_t, t)
+ \lambda \sigma_t \epsilon_t, \\
\hat{\Psi}^R_0 &= 
\frac{1}{\sqrt{\bar{\alpha}_t}}
\left(\Psi^R_t - \sqrt{1-\bar{\alpha}_t} \epsilon_\theta(\mathcal{G}_t, t)\right),
\label{eq:9}
\end{align}
\vspace{-3mm}

where $t, t' \in \mathcal{T}^S$ are diffusion timesteps with $t' < t$, 
$\epsilon_\theta(\mathcal{G}_t, t)$ denotes the predicted noise from the denoiser, 
$\sigma_t = \sqrt{(1-\bar{\alpha}_{t'})/(1-\bar{\alpha}_{t})} \sqrt{1-\bar{\alpha}_{t}/\bar{\alpha}_{t'}}$ 
is the variance term defined in~\cite{song2020denoising}, 
and $\lambda$ controls the level of random noise at each inference step. Once $\Psi^R_t$ is denoised to $\Psi^R_{t'}$, link nodes and all edges are generated according to Eq.~\ref{eq:7} to maintain node–edge spatial consistency. Through this iterative process, a complete noisy $\mathcal{T(R,O)}$ Graph can be progressively denoised into a clean graph, which represents a valid grasp.

\subsection{Training and Inference}
\label{sec:3.3}
$\mathcal{T(R,O)}$ Grasp supports end-to-end training, allowing the model to learn spatial transformations across multiple object-embodiment pairs. During inference, $\mathcal{T(R,O)}$ enables grasp synthesis either unconditionally or conditioned on an initial hand status. In contrast to $\mathcal{D(R,O)}$~\cite{wei2024dro} Grasp, it removes the reliance on an initial reference hand status, thereby preventing any suboptimal or infeasible initial poses from degrading grasp quality. Moreover, $\mathcal{T(R,O)}$ allows dynamic adjustment of DDIM inference steps and incorporates the GPU-based PyroKi~\cite{pyroki2025} optimization, resulting in a much more light-weighted and efficient inference process.

\noindent \textbf{Loss Function.} The training objective of diffusion models is to maximize the data log-likelihood via a variational lower bound, which can be simplified to minimize the noise prediction error according to~\cite{ho2020denoising}. Therefore, the training loss of the denoising model can be simplified as:

\vspace{-5mm}
\begin{equation}
L := \gamma_{p}||\epsilon^{p} -\epsilon_{\theta}^{p}(\mathcal{G}_t, t)||^2 + \gamma_{r}||\epsilon^{r} -\epsilon_{\theta}^{r}(\mathcal{G}_t, t)||^2,
\label{eq:10}
\end{equation}
\vspace{-5mm}

where $\epsilon^p, \epsilon^r$ represent position and rotation noise on link poses, and $\gamma_p, \gamma_r$ denote corresponding loss weight.

\noindent \textbf{Unconditioned Inference.} Benefiting from its diffusion-based structure, $\mathcal{T(R,O)}$ can generate precise grasps without relying on an initial pose, thereby avoiding the risk of poor initialization degrading grasp quality as in $\mathcal{D(R,O)}$. For efficient and adaptive inference, we adopt the DDIM~\cite{song2020denoising} sampling scheme. We sample $M$ diffusion timestamps $\mathcal{T}^S=\{t_i\}_{i=1}^M$ from the full schedule $\{1,\dots,T\}$ and apply the reverse process as described in Sec.~\ref{sec:3.2}.

\noindent \textbf{Conditioned Inference.} Grasp synthesis under specific conditions are important in real-world deployment. Based on guidance strategies in diffusion inference~\cite{Ho2022ClassifierFreeDG, bansal2023universal, guo2024gradient, meng2022sdedit}, $\mathcal{T(R,O)}$ naturally supports grasp synthesis conditioned on diverse grasp configurations, such as initial status and contact points. For example, given an initial hand status, we first inject Gaussian noise $\epsilon$ to the initial link poses to an intermediate timestamp $t^{\star}$ corresponding to the noise level $\bar{\alpha}_{t^{\star}}$ as Eq.~\ref{eq:11}, where $\Psi^R_{\text{init}}$ denotes link nodes initialized from given hand status. 

\vspace{-3mm}
\begin{equation}
\Psi^{R}_{t^\star, \text{init}} 
= \sqrt{\bar{\alpha}_{t^\star}} \, \Psi^R_{\text{init}}
  + \sqrt{1-\bar{\alpha}_{t^\star}} \, \epsilon,
\label{eq:11}
\end{equation}
\vspace{-5mm}

Starting from the intermediate status $\Psi^{R, \text{init}}_{t^\star} $, we employ the gradient of geodesic distance on $\text{SO(3)}$ to guide the predicted palm orientation toward the palm orientation of the initial status $R_{\text{init}} \in \text{SO}(3)$:

\vspace{-5mm}
\begin{align}
L_{\text{geo}}(R_1, R_2) = \text{arccos}(\frac{\text{tr}(R_{1}R_{2}^T)-1}{2}), \\
\hat{\Psi}^R_0 \leftarrow \hat{\Psi}^R_0 - s(t)\nabla_{\hat{\Psi}^R_0} L_{\text{geo}}(R_{\text{init}}, \hat{\Psi}^R_{0,\text{palm}})), \\
\epsilon_{\theta}(\mathcal{G}_t, t) = \frac{1}{ \sqrt{1-\bar{\alpha}_t}}(\Psi_t^R -\sqrt{\bar{\alpha}_t} \hat{\Psi}^R_0),
\label{eq:12}
\end{align}
\vspace{-4mm}

where $s(t) \in [0, g_s]$ controls the guidance strength for each sampling step and $\hat{\Psi}^R_{0,\text{palm}}$ denotes the palm rotation matrix extracted from the denoised estimation of link poses at the current timestep. Through gradient-based guidance, the generated grasps are encouraged to follow the initial hand status, while ensuring that the hand–object transformations correspond to a reliable grasp. Notably, such conditioned grasp synthesis further empowers $\mathcal{T(R,O)}$ Grasp to support closed-loop manipulation in dynamic scenarios.

\noindent \textbf{Pyroki Joint Optimization.} After obtaining the clean link poses for a potential grasp, we formulate a inverse kinematics problem to compute the joint values $\boldsymbol{q}$ that best realize these predicted poses:

\vspace{-5mm}
\begin{align}
\min_{\boldsymbol{q}} \ \sum_{i=1}^{L}
|| \text{FK}_{i}(\boldsymbol{q}) - \psi_i^R||_2,
\quad
\text{s.t. } \boldsymbol{q} \in [\boldsymbol{q}_{min}, \boldsymbol{q}_{max}].
\label{eq:13}
\end{align}
\vspace{-3mm}

where $\text{FK}_i(\cdot)$ and $\psi_i^R$ denote the forward kinematics and predicted pose of link $i$, respectively, and $[\boldsymbol{q}_{min}, \boldsymbol{q}_{max}]$ denotes the joint limits. To achieve efficient IK, we integrate Pyroki~\cite{pyroki2025}, a highly efficient, modular, and extensive toolkit for solving kinematics optimization problems, into our inference pipeline. Once the scripts are pre-compiled in JAX~\cite{jax2018github}, IK process with Pyroki toolkit can achieve real-time performance at over 30 FPS on our tested dexterous hands (i.e., Allegro, Barrett and Shadowhand).
%%%%%%%%%%%%%%%%%%%%%%%%%%%%%%%%%%%%%%%%%%%%%%%%%%%%%%%%%%%%%%%%%%%%%%%%%%%%%%%%
\vspace{5pt}
\section{Experiments}
In this section, we perform a series of experiments aimed
at addressing the following questions (Q1-Q7):
\begin{itemize}[leftmargin=5pt]
    \item[] Q1: How successful and diverse is the $\mathcal{T(R,O)}$ unconditioned grasp synthesis? (Sec.~\ref{sec:4.2})
    \item[] Q2: How does the model leverage the guidance from the initial pose and contact regions to perform conditioned grasp synthesis? (Sec.~\ref{sec:4.3})
    \item[] Q3: Is $\mathcal{T(R,O)}$ representation generalizable across multiple dexterous hand embodiments? (Sec.~\ref{sec:4.2})
    \item[] Q4: Is $\mathcal{T(R,O)}$ representation robust to partial and incomplete object observations? (Sec.~\ref{sec:4.2})
    \item[] Q5: How computationally efficient is $\mathcal{T(R,O)}$ Grasp during training and inference? (Sec.~\ref{sec:4.4})
    \item[] Q6: Does $\mathcal{T(R,O)}$ Grasp support closed-loop dexterous manipulation in dynamic environments? (Sec.~\ref{sec:4.5})
%    \item[] Q7: Can $\mathcal{T(R,O)}$ Grasp zero-shot to novel hand embodiments with comparable geometries? (Sec.~\ref{sec:4.6})
    \item[] Q7: How does $\mathcal{T(R,O)}$ perform in real-world? (Sec.~\ref{sec:4.7})
\end{itemize}

%In this section, we comprehensively evaluate the performance and efficiency of $\mathcal{T(R,O)}$ Grasp in both unconditioned and conditioned grasp synthesis across multiple dexterous hand embodiments. Under various settings, $\mathcal{T(R,O)}$ achieves \textbf{state-of-the-art} grasping success rate, while offering an average of $\mathbf{3\times}$ \textbf{faster} inference speed and $\mathbf{8\times}$ FPS compared to $\mathcal{D(R,O)}$ Grasp~\cite{wei2024dro}.

\begin{table*}[htbp]
    \centering
    \renewcommand\arraystretch{1.2}
    \captionsetup{justification=raggedright, singlelinecheck=false}
    \resizebox{\textwidth}{!}{
        \begin{threeparttable}
            \begin{tabular}{c|ccc|c|ccc|ccc}
                \toprule
                \multirow{2}{*} {\textbf{Method}} 
                & \multicolumn{4}{c|}{\textbf{Success Rate (\%) $\uparrow$}}
                & \multicolumn{3}{c|}{\textbf{Efficiency (sec.) $\downarrow$}}
                & \multicolumn{3}{c}{\textbf{Diversity (rad.) $\uparrow$}}
                \\ 
                \cline{2-11} 
                & Barrett & Allegro & ShadowHand & Avg.
                & Barrett & Allegro & ShadowHand
                & Barrett & Allegro & ShadowHand
                \\ \hline
                DFC~\cite{dfc}
                    & 86.30 & 76.21 & 58.80 & 73.77
                    & $>$1800 & $>$1800 & $>$1800
                    & 0.532 & \textbf{0.454} & 0.435
                \\
                GenDexGrasp~\cite{li2023gendexgrasp}
                    & 67.00 & 51.00 & 54.20 & 57.40
                    & 14.67 & 25.10 & 19.34
                    & 0.488 & 0.389 & 0.318
                \\
                ManiFM~\cite{xu2024manifoundation}
                    & - & 42.60 & - & 42.60
                    & - & 9.07 & -
                    & - & 0.288 & -
                \\
                DRO-Grasp~\cite{wei2024dro}
                    & 87.30 & 92.30 & 83.00 & 87.53
                    & 0.49 & 0.47 & 0.98
                    & 0.513 & 0.397 & \textbf{0.441}
                \\
                \textbf{TRO-Grasp (cond.)}
                    & 89.80 & 93.70 & 93.50 & 92.33
                    & \textbf{0.19} & \textbf{0.21} & \textbf{0.22}
                    & \textbf{0.561} & 0.430 & 0.326
                \\
                \textbf{TRO-Grasp (uncond.)}
                    & \textbf{91.90} & \textbf{94.00} & \textbf{98.60} & \textbf{94.83}
                    & 0.23 & 0.23 & 0.25
                    & 0.542 & 0.370 & 0.292
                \\
                % \textbf{TRO-Grasp (conditioned)}
                %     & 89.80 & 93.70 & 93.50 & 92.33
                %     & \textbf{0.17} & \textbf{0.18} & \textbf{0.19}
                %     & \textbf{0.561} & 0.430 & 0.326
                % \\
                % \textbf{TRO-Grasp (unconditioned)}
                %     & \textbf{91.90} & \textbf{94.00} & \textbf{98.60} & \textbf{94.83}
                %     & 0.19 & 0.20 & 0.22
                %     & 0.542 & 0.370 & 0.292
                % \\
                \bottomrule
            \end{tabular}
        \end{threeparttable}
    }
    \caption{Overall performance and inference efficiency comparison with baselines: For ManiFM~\cite{xu2024manifoundation}, evaluation is limited to the available Allegro hand checkpoint.}
    \label{tab:main result}
    \vspace{-5pt}
\end{table*}

\begin{table}[t]
    \centering
    \renewcommand\arraystretch{1.2}
    \captionsetup{justification=raggedright, singlelinecheck=false}
    \resizebox{\columnwidth}{!}{
        \begin{tabular}{c|ccc|ccc}
            \toprule
            \multirow{2}{*} {\textbf{Method}} 
            & \multicolumn{3}{c|}{\textbf{Success Rate (\%) $\uparrow$}} 
            & \multicolumn{3}{c}{\textbf{Diversity (rad) $\uparrow$}}
            \\
            \cline{2-7} 
            & Barrett & Allegro & ShadowHand 
            & Barrett & Allegro & ShadowHand
            \\ 
            \hline
            Single
                & 91.60 & 92.30 & 96.00
                & 0.528 & 0.372 & 0.281
            \\
            Multi
                & \textbf{91.90} & \textbf{94.00} & \textbf{98.60}
                & 0.542 & 0.370 & \textbf{0.292}
            \\ 
            Partial
                & 86.80 & 93.40 & 94.60
                & \textbf{0.548} & \textbf{0.391} & 0.290
            \\ 
            \bottomrule
        \end{tabular}
    }
    \caption{Robust and generalizable performance of $\mathcal{T(R,O)}$: Single, Multi, and Partial denote training and evaluation on single-hand, multi-hand, and partial object point cloud datasets, respectively.}
    \label{tab:cross and partial}
    \vspace{-10pt}
\end{table}

\subsection{Experimental Setting}
\label{sec:4.1}

\noindent \textbf{Dataset.} Following $\mathcal{D(R,O)}$ Grasp, we evaluate on a filtered subset of CMapDataset~\cite{li2022gendexgrasp}, which includes dexterous hand embodiments with 3–5 fingers (Barrett, Allegro and ShadowHand). The training dataset consists of 14,011 grasp demonstrations on 48 objects, while the testing dataset evaluates 100 grasps per object across 10 unseen objects. For every object, we sample 65,536 points from the mesh file, and randomly sample 512 points in each training iteration.

\noindent \textbf{Evaluation Metrics.} Similar to $\mathcal{D(R,O)}$ Grasp, we report the following three metrics:
\begin{itemize}[leftmargin=*]
    \item \textit{Success Rate}: a grasp is considered successful in Isaac Gym if the object displacement remains below 2\,cm under six directional perturbations.  
    \item \textit{Efficiency}: measured as the computational time per grasp, including both DDIM inference and Pyroki IK process. 
    \item \textit{Diversity}: quantified as the standard deviation of joint values across all successful grasps.
\end{itemize}

\noindent \textbf{Implementation Details.} For $\mathcal{T(R,O)}$ Graph construction, each object is encoded into $P=25$ nodes, and link nodes are zero-padded to $L=25$ across all embodiments to enable parallel computing. In the $\mathcal{T(R,O)}$ Graph Diffusion process, we adopt a linear scheduler for the noise variance and set the total number of diffusion steps to $T=1,000$. During inference, we sample $M=20$ steps and set $\lambda=0.2$ to inject random noise for more diverse grasp synthesis. For the loss function, we set $\gamma_p=\gamma_r=1.0$ and apply a sine-based guidance strength $s(t)$ for conditioned grasp synthesis. All $\mathcal{T(R,O)}$ models are trained for 300 epochs on an NVIDIA A100 40GB GPU with Adam optimizer and step learning rate scheduler. Additional implementation details and parameter settings are provided in the Appendix.

\subsection{Unconditioned grasp synthesis}
\label{sec:4.2}

\begin{figure}[t] \centering
    \includegraphics[width=0.98 \linewidth]{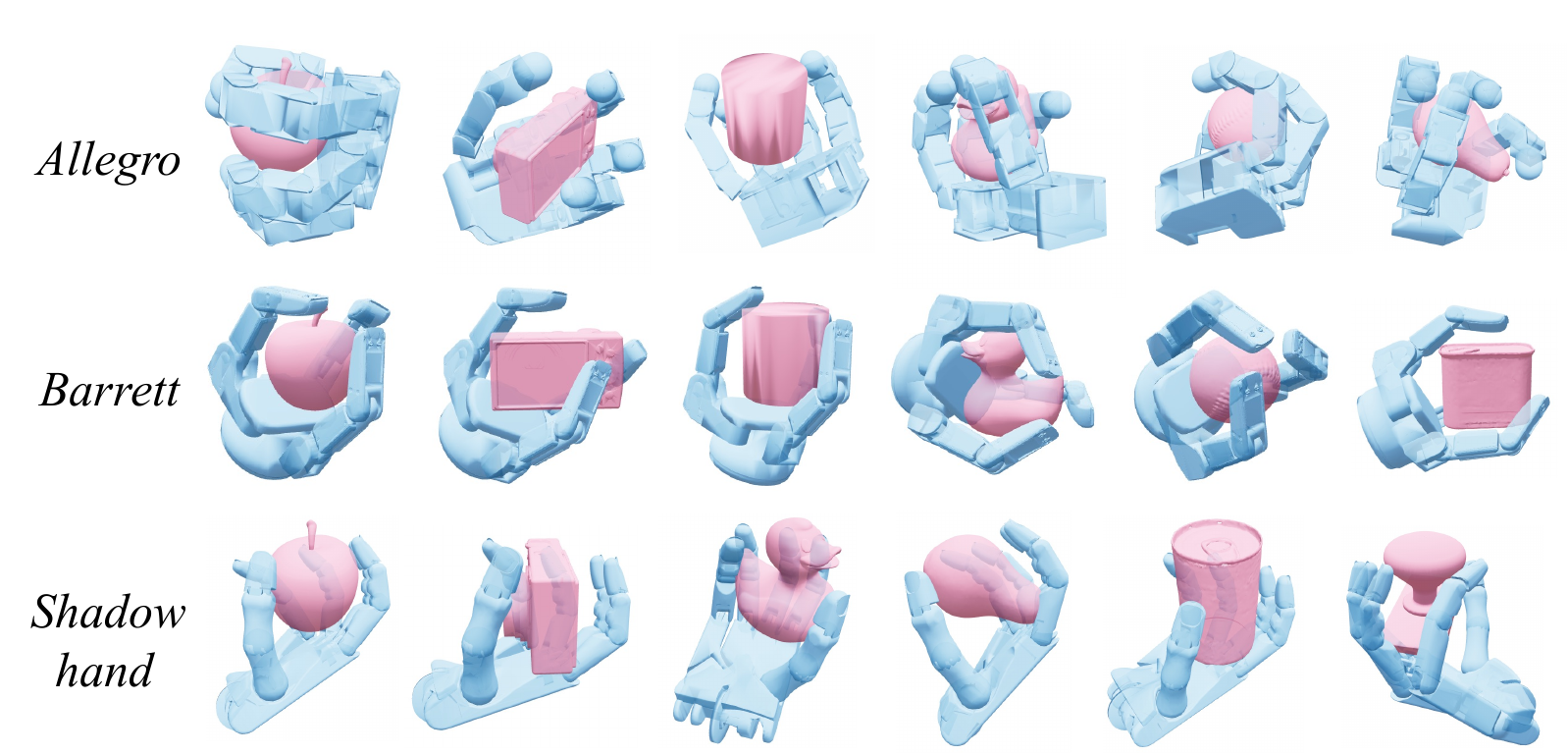}
    \caption{Diverse and accurate unconditioned grasp synthesis of $\mathcal{T(R,O)}$ across multiple dexterous hand embodiments.}
    \vspace{-5pt}
    \label{fig:unconditonal grasp}
\end{figure}

\begin{figure}[t] \centering
    \includegraphics[width=0.98 \linewidth]{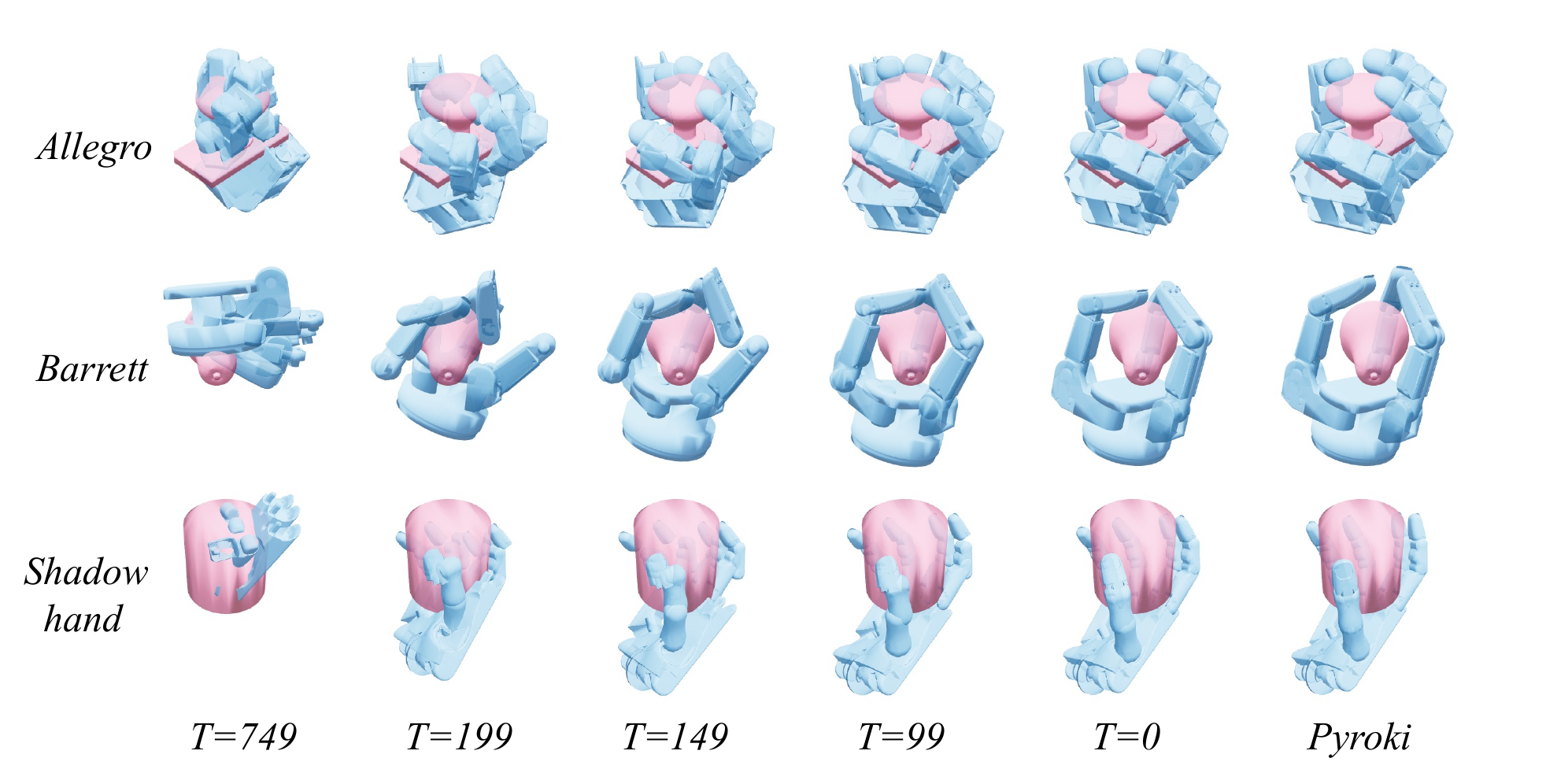}
    \caption{Visualization of unconditioned DDIM inference across timesteps. ``Pyroki'' indicates the grasp after IK with the Pyroki~\cite{pyroki2025} toolkit.}
    \vspace{-15pt}
    \label{fig:diffusion process}
\end{figure}

\noindent \textbf{Overall Performance.} As illustrated in Tab.~\ref{tab:main result}, our method substantially outperforms all baselines in grasping success rate, achieving an average increase of 7.3\% over $\mathcal{D(R,O)}$ and up to 15.6\% on ShadowHand. Fig.~\ref{fig:unconditonal grasp} shows diverse and accurate grasp on unseen objects across Allegro, Barrett and Shadowhand. In Fig.~\ref{fig:diffusion process}, we further visualize the diffusion inference process of $\mathcal{T(R,O)}$, where noisy link poses gradually converge to form a valid grasp. In addition, $\mathcal{T(R,O)}$ accelerates inference by $3\times$ in average compared to $\mathcal{D(R,O)}$ Grasp, owing to more efficient $\mathcal{T(R,O)}$ representation and DDIM-based diffusion inference, along with the real-time Pyroki IK toolkit.

\noindent \textbf{Cross-embodiment Generalizability.} To validate the generalizability of the $\mathcal{T(R,O)}$ representation, we conduct grasp synthesis across multiple embodiments. As shown in Tab.~\ref{tab:cross and partial}, $\mathcal{T(R,O)}$ Grasp achieves superior performance through multi-embodiment training, demonstrating that the $\mathcal{T(R,O)}$ Graph representation, together with the graph diffusion model, generalizes effectively across diverse dexterous hands. These results further suggests the feasibility of scaling $\mathcal{T(R,O)}$ Grasp into a large-scale foundation model for dexterous grasping by training it on extensive multi-hand grasp demonstrations.

\noindent \textbf{Partial Observation.} In practical scenarios, object observations are often incomplete due to occlusions or limited viewpoints. To assess the robustness of $\mathcal{T(R,O)}$ Grasp under such conditions, we conduct both training and testing with one-side $50\%$ point cloud inputs for all objects. As shown in Tab.~\ref{tab:cross and partial}, $\mathcal{T(R,O)}$ Grasp maintains high performance despite partial observations, demonstrating its robustness for real-world deployment with incomplete observations.

\subsection{Conditioned grasp synthesis}
\label{sec:4.3}

\begin{figure}[t] \centering
    \includegraphics[width=0.98 \linewidth]{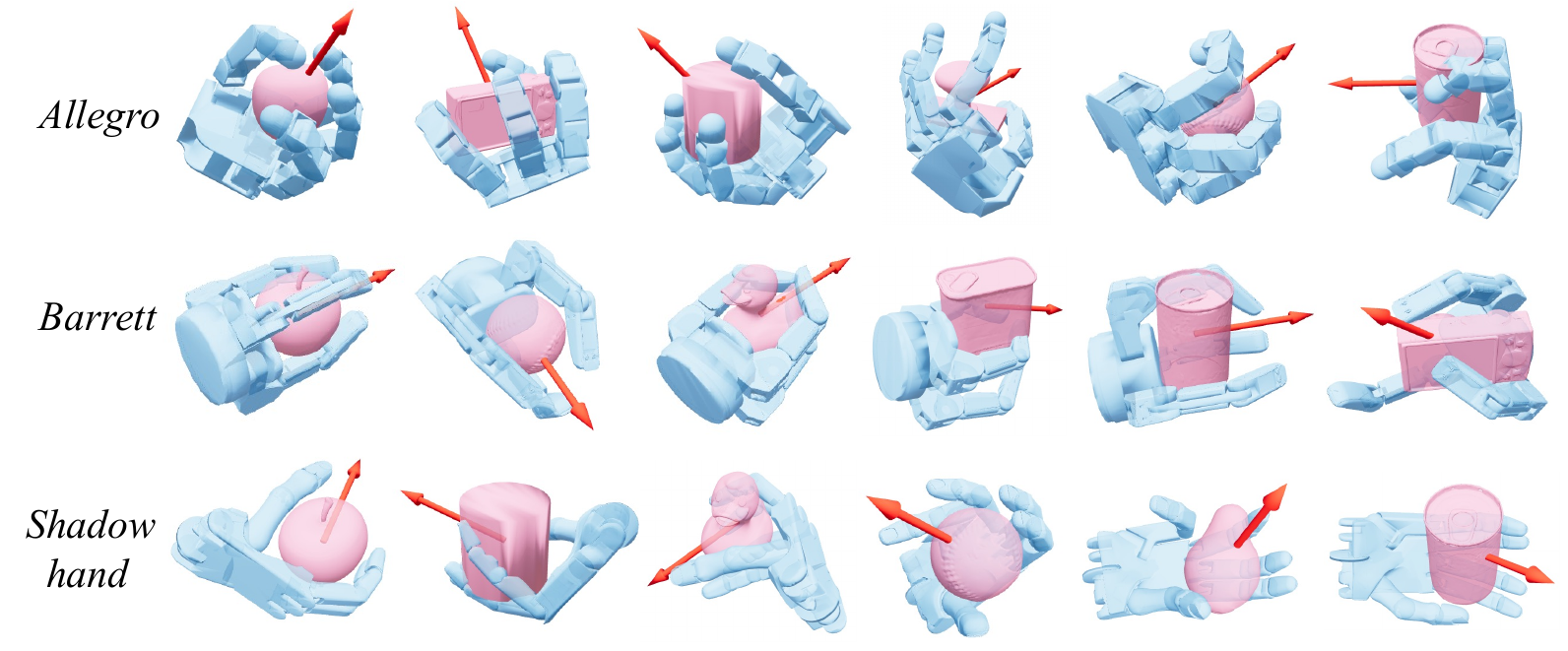}
    \caption{grasp synthesis conditioned on initial hand status: Red arrow denotes the palm direction of pose guidance.}
    \vspace{-2pt}
    \label{fig:conditonal grasp}
\end{figure}

Conditioned grasp synthesis is crucial for real-world deployment scenarios, such as grasping objects from a table or retrieving them from a shelf. In this experiment, we incorporate gradient-based guidance into $\mathcal{T(R,O)}$ Grasp, enabling grasp synthesis conditioned on a given initial hand status or contact region. 

\noindent \textbf{Grasp Synthesis Conditioned on Initial Pose.} As indicated in Tab.~\ref{tab:main result}, grasp synthesis conditioned on given initial pose trades a slight drop in success rate for higher grasp diversity and faster inference. Visualization in Fig.~\ref{fig:conditonal grasp} demonstrates that the generated grasps align well with the palm directions of given hand status. A diverse set of initialization grasping poses increases the risk of generating infeasible grasps, while simultaneously promoting grasp diversity. In addition, starting the diffusion inference process at an intermediate timestep enables faster inference at approximately 5 FPS, indicating the potential of incorporating $\mathcal{T(R,O)}$ framework into closed-loop dexterous hand manipulation (Sec.~\ref{sec:4.5}).

\noindent \textbf{Grasp Synthesis Conditioned on Contact Region.} Similarly, $\mathcal{T(R,O)}$ Grasp enables grasp synthesis conditioned on a specified contact region. Given the object point cloud $P^O \in \mathbb{R}^{N \times 3}$, we randomly sample a contiguous subset of $K$ points, denoted as $P^O_{\text{cont}} \subseteq P^O$ to represent the contact region. Each point within the contact region is assigned a Gaussian heat value according to its distance from the contact center, resulting in a contact heat map $H = \{h_i\}_{i=1}^K$. Then, we define the contact loss $L_{\text{cont}}$ as Eq.~\ref{eq:contact}. $R_{\text{cont}}$ represents the palm pose whose normal direction opposes the object normal at the contact center. $L_{\text{geo}}$ denotes geodesic distance on $\text{SO(3)}$ and $L_{\text{dis}}$ denotes the weighted distance between contact points and link centers. 

\vspace{-13pt}
\begin{equation}
\label{eq:contact}
L_{\text{cont}} = L_{\text{geo}}(\hat{\Psi}_{0, \text{palm}}^R, R_{\text{cont}}) + L_{\text{dis}}(H, P^O_{\text{cont}}, \hat{\Psi}_{0}^R).
\end{equation}

\begin{figure}[t] \centering
    \includegraphics[width=0.98 \linewidth]{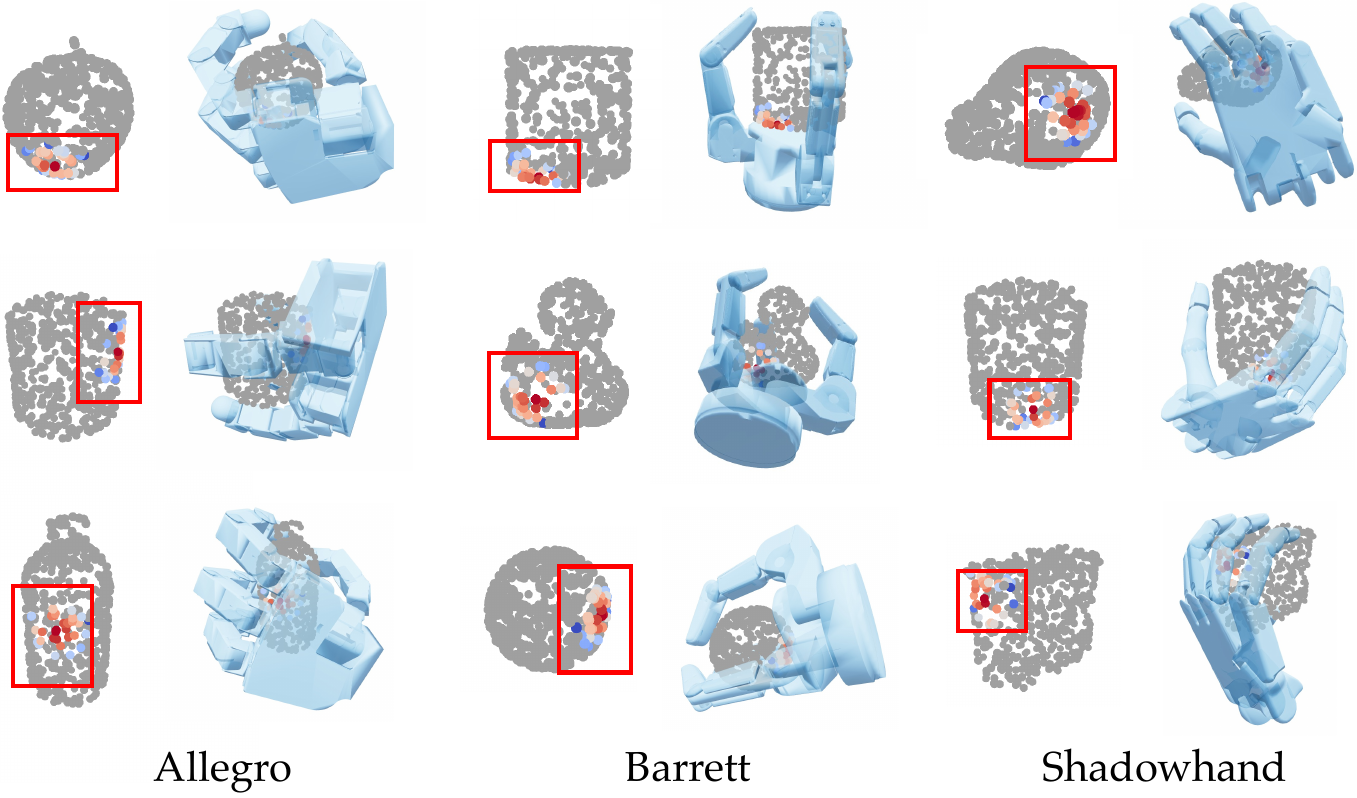}
    \caption{Grasp synthesis conditioned on contact regions. Red box denotes the contact heat map guidance.}
    \vspace{-5pt}
    \label{fig:contact grasp}
\end{figure}

As described in Sec.~\ref{sec:3.3}, the gradient of the contact loss is applied during the diffusion inference process. As shown in Fig.~\ref{fig:contact grasp} and Tab.~\ref{tab:contact-guidance}, $\mathcal{T(R,O)}$ Grasp achieves promising results with high diversity on grasp synthesis conditioned on given contact regions.

\begin{table}[t]
    \centering
    \renewcommand{\arraystretch}{1.15}
    \captionsetup{justification=centering, singlelinecheck=false}
    \resizebox{\columnwidth}{!}{
    \begin{tabular}{ccc@{\hspace{1em}}ccc}
        \toprule
        \multicolumn{3}{c}{\textbf{Success Rate (\%) $\uparrow$}} &
        \multicolumn{3}{c}{\textbf{Diversity (rad) $\uparrow$}} \\
        \midrule
        \textbf{Barrett} & \textbf{Allegro} & \textbf{ShadowHand} &
        \textbf{Barrett} & \textbf{Allegro} & \textbf{ShadowHand} \\
        \midrule
        83.6 & 91.0 & 87.8 & 0.559 & 0.482 & 0.331 \\
        \bottomrule
    \end{tabular}
    }
    \caption{Contact-conditioned grasp synthesis.}
    \label{tab:contact-guidance}
    \vspace{-15pt}
\end{table}

\subsection{Efficiency Analysis}
\label{sec:4.4}

In this section, we provide a comprehensive efficiency analysis of $\mathcal{T(R,O)}$ Grasp in both training and inference on an NVIDIA A100 40GB GPU. To evaluate memory usage, we fix the batch size to 4 in training and 25 in inference, and report the average memory usage per sample for fair comparison. To evaluate throughput, we increase the batch size of all models to the maximum that fits into 40GB memory. As indicated in Tab.~\ref{tab:efficiency}, $\mathcal{T(R,O)}$ requires only $43\%$ training memory and $21\%$ inference memory compared to $\mathcal{D(R,O)}$, while possessing $2\times$ more model parameters with stronger representation capability. In addition, $\mathcal{T(R,O)}$ demonstrates much higher throughput and inference speed owing to its lightweight architecture, DDIM-based inference strategy, and Pyroki-based IK optimization. As a result, it can be deployed on real-world robotic platforms with limited computational resources.

\begin{table}[t]
    \centering
    \renewcommand\arraystretch{1.2}
    \captionsetup{justification=raggedright, singlelinecheck=false}
    \resizebox{\columnwidth}{!}{
        \begin{tabular}{c|ccc}
            \toprule
            \textbf{Method} 
            & DRO
            & TRO (cond.)
            & TRO (uncond.)
            \\
            \hline
            Infer Mem. (GB/sample) & 0.38 & 0.08 & 0.08 \\
            Infer throughput (sample/s) & 5 & 50 & 32 \\
            Model Param. (M) & 14.00 & 28.06 & 28.06 \\
            Train Mem. (GB/sample) & 4.10 & 1.77 & 1.77 \\
            Train throughput (sample/s) & 7 & 22 & 22 \\
            \bottomrule
        \end{tabular}
    }
    \caption{Training and Inference Efficiency Comparison.}
    \label{tab:efficiency}
    \vspace{-10pt}
\end{table}

\begin{table}[t]
    \centering
    \renewcommand\arraystretch{1.2}
    \captionsetup{justification=raggedright, singlelinecheck=false}
    \resizebox{\linewidth}{!}{
        \begin{tabular}{cccccccccc}
            \toprule
            \multicolumn{2}{c}{\textbf{Apple}} & \multicolumn{2}{c}{\textbf{Camera}} & \multicolumn{2}{c}{\textbf{Cylinder}} & \multicolumn{2}{c}{\textbf{Door Knob}} & \multicolumn{2}{c}{\textbf{Rubber Duck}} \\
            Const. & Rand. & Const. & Rand. & Const. & Rand. & Const. & Rand. & Const. & Rand. \\
            100.0 & 100.0 & 76.7 & 93.3 & 93.3 & 96.7 & 96.7 & 90.0 & 96.7 & 93.3 \\
            \midrule
            \multicolumn{2}{c}{\textbf{Water Bottle}} & \multicolumn{2}{c}{\textbf{Baseball}} & \multicolumn{2}{c}{\textbf{Pear}} & \multicolumn{2}{c}{\textbf{Meat Can}} & \multicolumn{2}{c}{\textbf{Soup Can}} \\
            Const. & Rand. & Const. & Rand. & Const. & Rand. & Const. & Rand. & Const. & Rand. \\
            56.7 & 70.0 & 100.0 & 96.7 & 93.3 & 83.3 & 93.3 & 90.0 & 86.7 & 96.7 \\
            \bottomrule
        \end{tabular}
    }
    \caption{Success rate of closed-loop grasp synthesis of Allegro hand on 10 unseen objects.}
    \label{tab:closed loop}
\end{table}

\subsection{Closed-Loop Grasping Evaluation}
\label{sec:4.5}

By leveraging its fast inference speed and the ability to perform grasp synthesis conditioned on previous grasp poses, $\mathcal{T(R,O)}$ Grasp achieves conditioned synthesis at an average of 5 FPS (Sec.~\ref{sec:4.3}), making closed-loop grasping feasible. Specifically, given a discrete time sequence $\{0, T, \dots, nT\}$, $\mathcal{T(R,O)}$ uses the pose from the previous timestamp $\Psi_{(i-1)T}^R$ as guidance to predict all link poses at the current timestamp $\Psi_{iT}^R$. We conduct closed-loop grasping evaluation with the Allegro hand on 10 unseen objects. The time interval is set to 0.25s with a total of 30 evaluation steps. At each interval, the object either translates at a constant velocity (5 cm/s) or undergoes random perturbations involve translational noise of up to 1.25 cm and rotational noise up to 30$^\circ$ per step. As shown in Tab.~\ref{tab:closed loop}, closed-loop $\mathcal{T(R,O)}$ Grasp achieves average success rates of $89.3\%$ under constant-velocity motion and $91.0\%$ under random perturbations, while open-loop approaches completely fail (0\% success rate). These results demonstrates the capability of $\mathcal{T(R,O)}$ Grasp as a general framework for closed-loop dexterous grasping across diverse dynamic scenarios.

\subsection{Real-Robot Experiments}
\label{sec:4.7}

\begin{figure}[t]
    \centering
    \includegraphics[width=\linewidth]{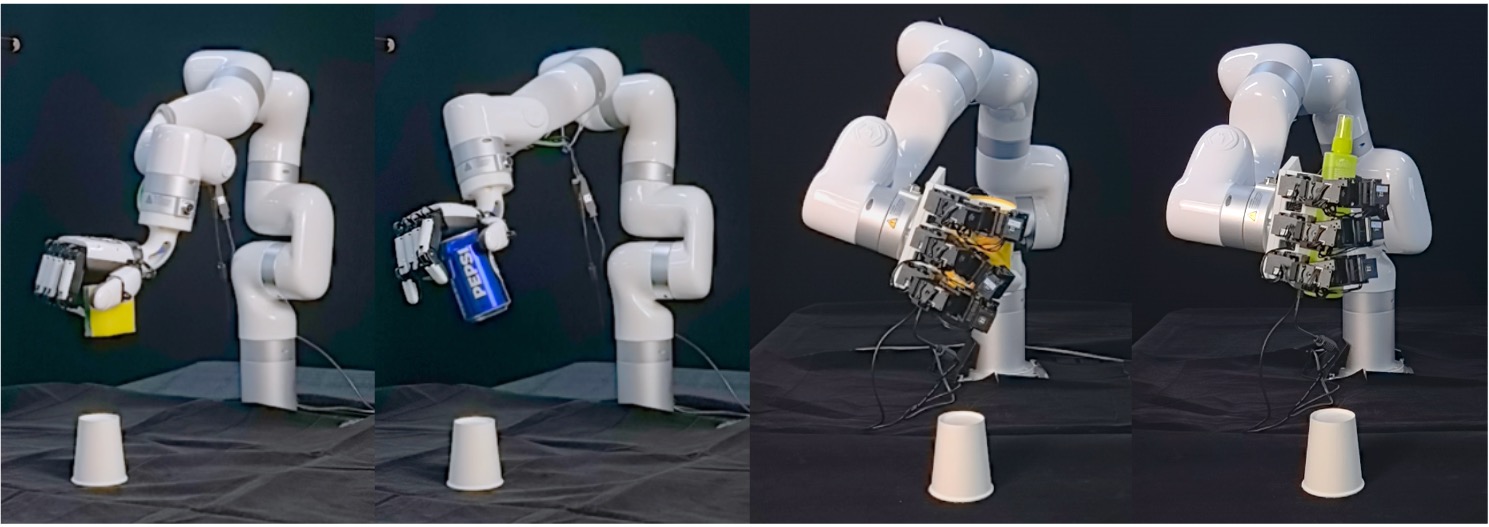}
    \vspace{-15pt} 
    \caption{Real-world experiment setting.}
    \vspace{-15pt}
    \label{fig:grasp_rw}
\end{figure}

We conduct real-world experiments with a uFactory xArm7 robot equipped with a Robotera XHand and LEAP Hand (as illustrated in Fig.~\ref{fig:grasp_rw}). Object point cloud observations are obtained using an overhead Intel RealSense D435 camera. We train our model on 7,800 grasp demonstrations and evaluate 100 grasps on 10 previously unseen objects for each dexterous hand. During inference, we leverage FoundationPose~\cite{wen2024foundationpose} to track the object pose. $\mathcal{T(R,O)}$ Grasp achieves an average success rate of 91\% and 90\% on XHand and LEAP Hand respectively, while enabling closed-loop grasping in dynamic environments. Additional experimental details, results, and videos are available in the appendix and on the project website~\href{https://nus-lins-lab.github.io/trograspweb/}{https://nus-lins-lab.github.io/trograspweb/}.

%%%%%%%%%%%%%%%%%%%%%%%%%%%%%%%%%%%%%%%%%%%%%%%%%%%%%%%%%%%%%%%%%%%%%%%%%%%%%%%%
\vspace{5pt}
\section{Conclusion}
\label{sec:5}

In this paper, we tackled the challenge of efficient and generalizable dexterous grasp synthesis. We highlighted the limitations of robot-centric, object-centric, and interaction-centric representations, and introduced the $\mathcal{T(R,O)}$ Graph—a unified, memory-efficient representation of object–robot interactions that generalizes across embodiments. Built upon this representation, we proposed a transformer-based graph diffusion model that enables both conditioned and unconditioned grasp generation, as well as closed-loop manipulation. Extensive experiments in simulation and on hardware show that $\mathcal{T(R,O)}$ Grasp outperforms all baselines on success rate with high efficiency and closed-loop manipulation. This work not only advances grasping methods but also lays the groundwork for large-scale dexterous grasping foundation models, with future directions in broader manipulation tasks, multimodal feedback, and large-scale pretraining.
%%%%%%%%%%%%%%%%%%%%%%%%%%%%%%%%%%%%%%%%%%%%%%%%%%%%%%%%%%%%%%%%%%%%%%%%%%%%%%%

%%%%%%%%%%%%%%%%%%%%%%%%%%%%%%%%%%%%%%%%%%%%%%%%%%%%%%%%%%%%%%%%%%%%%%%%%%%%%%%%

{\small
\bibliographystyle{IEEEtran}
\bibliography{main}
}

\clearpage
\appendix
\subsection{Real-World Experiment}
\label{sec:6.1}

\subsubsection{Data Collection and Model Training} To perform real-world dexterous grasping, we collect LEAP Hand Dataset and XHand Dataset to train $\mathcal{T(R,O)}$ Grasp models independently. We select 50 objects from ContactDB~\cite{brahmbhatt2019contactdb} and 28 objects from YCB dataset~\cite{calli2017yale}, applying DFC-based~\cite{dfc} grasp optimization to generate grasping demonstrations. After filtering, we obtain 7,800 grasp demonstrations for each hand as the training dataset. Following the training setup in Sec.~\ref{sec:4.1}, we train $\mathcal{T(R,O)}$ Grasp on the collected dataset and evaluate its performance in real-world scenarios.

\subsubsection{Real-world Deployment} First, we use AR Code~\cite{arcode2022} to scan 10 novel objects for each hand. After camera calibration, we employ FoundationPose~\cite{wen2024foundationpose} to estimate the object pose from monocular RGB-D input captured by an Intel RealSense D435 camera. The point cloud input for each object is then obtained by transforming the sampled point cloud from the scanned 3D model into the world frame. To avoid collision in the tabletop grasp setting, we randomly sample an initial hand pose from top-down to right-side orientations, while taking the sampled initial pose as guidance during $\mathcal{T(R,O)}$ grasp synthesis. Then, we use MPLib~\cite{MPlib2023} for xArm motion planning to reach the desired end-effector pose. For closed-loop grasping in dynamic environments, we place the object on a conveyor belt and employ FoundationPose tracking to continuously update its pose, repeating the above process in real time.

\subsubsection{Experiment Results} 
\begin{figure*}[htbp]
    \centering
    \includegraphics[width=\linewidth]{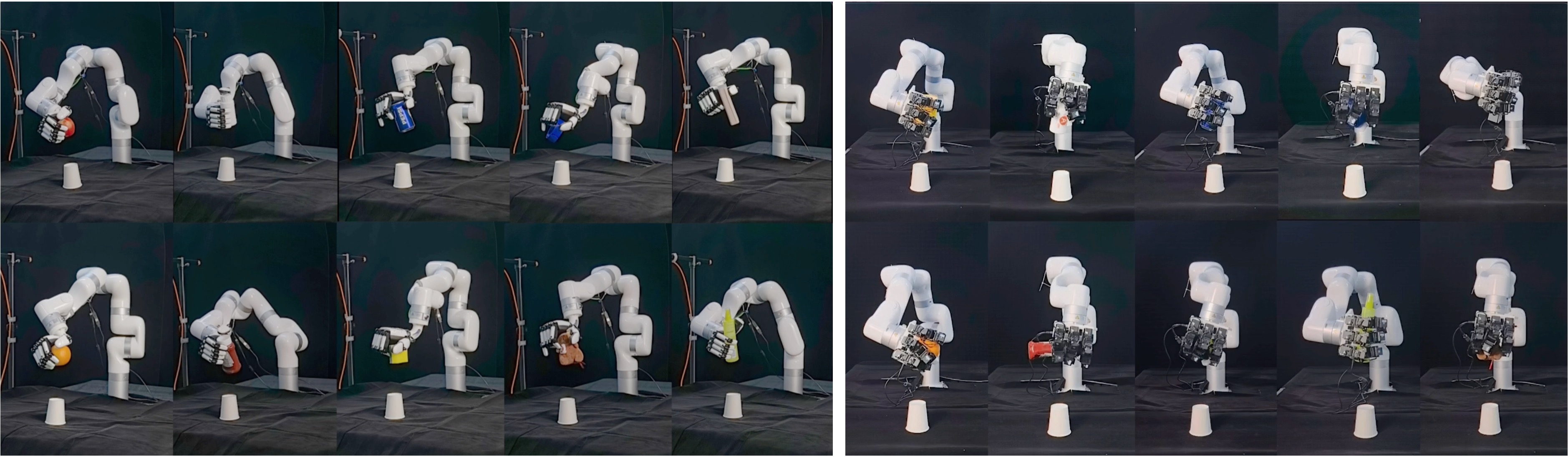}
    \caption{Real-world grasp synthesis on XHand (left) and LEAP Hand (right).}
    \vspace{-5pt}
    \label{fig:realworld}
\end{figure*}

\begin{figure*}[htbp]
    \centering
    \includegraphics[width=\linewidth]{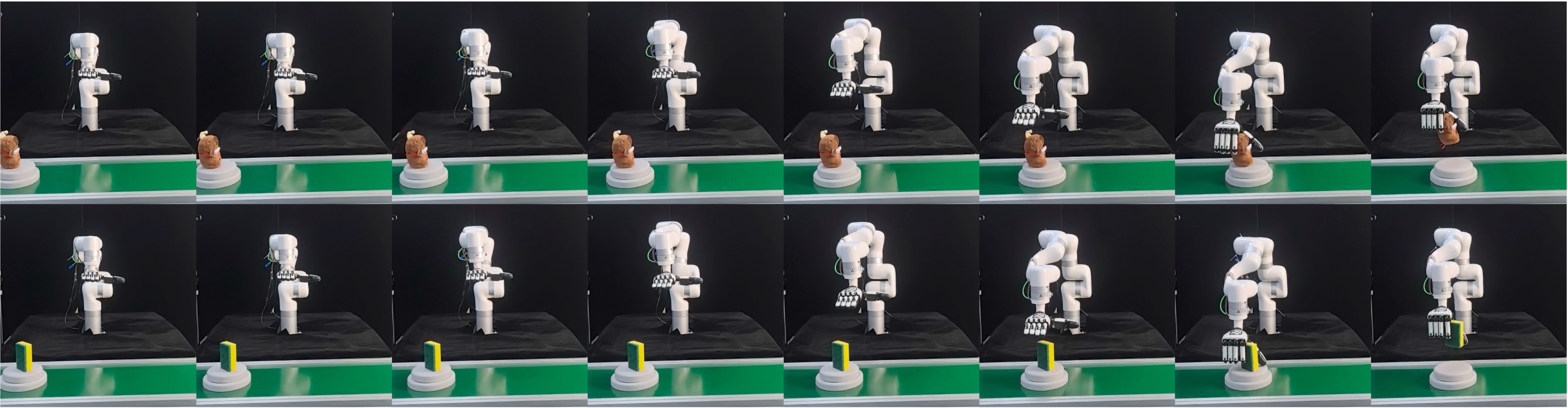}
    \caption{Real-world closed-loop grasp synthesis in dynamic environments.}
    \vspace{-5pt}
    \label{fig:realworld-dynamic}
\end{figure*}

As shown in Tab.~\ref{tab:realworld-xhand} and Tab.~\ref{tab:realworld-leaphand}, $\mathcal{T(R,O)}$ Grasp achieves an average success rate of \textbf{91\%} and \textbf{90\%} on XHand and LEAP Hand, respectively. Visualization in Fig.~\ref{fig:realworld} demonstrates that our method performs robust and generalizable grasp synthesis on novel objects. Furthermore, Fig.~\ref{fig:realworld-dynamic} indicates that the high inference speed of $\mathcal{T(R,O)}$ Grasp enables closed-loop grasp synthesis, allowing it to successfully capture moving objects on a conveyor belt. Complete videos of real-world experiments are available on the project website~\href{https://nus-lins-lab.github.io/trograspweb/}{https://nus-lins-lab.github.io/trograspweb/}.

\begin{table}[ht]
    \centering
    \small
    \renewcommand\arraystretch{1}
    \captionsetup{justification=centering, singlelinecheck=false}
    \resizebox{\linewidth}{!}{
        \begin{tabular}{ccccc}
            \toprule
            \textbf{Apple} & \textbf{Bottle} & \textbf{Cola} & \textbf{Cylinder} & \textbf{Box}
            \\
            9/10 & 10/10 & 9/10 & 8/10 & 9/10
            \\
            \midrule
            \textbf{Orange} & \textbf{Sauce} & \textbf{Sponge} & \textbf{Toy} & \textbf{Spray Bottle}
            \\
            10/10 & 8/10 & 9/10 & 10/10 & 9/10
            \\
            \bottomrule
        \end{tabular}
    }
    \caption{Real-world experiment results on XHand.}
    \label{tab:realworld-xhand}
    \vspace{-15pt}
\end{table}

\begin{table}[ht]
    \centering
    \small
    \renewcommand\arraystretch{1}
    \captionsetup{justification=centering, singlelinecheck=false}
    \resizebox{\linewidth}{!}{
        \begin{tabular}{ccccc}
            \toprule
            \textbf{Chip Box} & \textbf{Bottle} & \textbf{Cola} & \textbf{Cylinder} & \textbf{Box}
            \\
            9/10 & 10/10 & 10/10 & 7/10 & 9/10
            \\
            \midrule
            \textbf{Orange} & \textbf{Sauce} & \textbf{Sponge} & \textbf{Spray Bottle} & \textbf{Toy}
            \\
            8/10 & 8/10 & 9/10 & 10/10 & 10/10
            \\
            \bottomrule
        \end{tabular}
    }
    \caption{Real-world experiment results on LEAP Hand.}
    \label{tab:realworld-leaphand}
    \vspace{-15pt}
\end{table}

\subsection{Network Architecture}
\label{sec:6:2}

\subsubsection{VQ-VAE Encoder} In $\mathcal{T(R,O)}$ Graph, we leverage the pretrained VQ-VAE encoder from~\cite{wang2025puzzlefusionpp} to partition object patches and extract corresponding geometry tokens. The object point cloud is first normalized within a unit sphere, and then encoded by the pretrained encoder into $P=25$ local geometry features $\{f_i^O\}_{i=1}^P$ along with corresponding patch center coordinates $\{c_i^O\}_{i=1}^P$.

\subsubsection{BPS Encoder} Since the number of point cloud varies for each link, we employ Basis Point Set (BPS)~\cite{prokudin2019efficient} algorithm to encode each link point cloud into a fixed-length geometric feature. Point clouds of the dexterous hand with $L$ links are defined as $\{P_i^R\}_{i=1}^{L}$ in their respective local frames, where each link point cloud is $P_i^R = \{p_{i1}, \ldots, p_{in_i}\} \in \mathbb{R}^{n_i \times 3}$. First, we normalize all points into a unit sphere:

\vspace{-2mm}
\begin{equation}
    p_{ij} = \frac{p_{ij} - \frac{1}{n_i} \sum_{j} p_{ij}}{\text{max}_{j} ||p_{ij} - \frac{1}{n_i} \sum_{j} p_{ij}||}, \ \forall \ i, j.
\end{equation}

Next, we randomly sample $B=124$ points within the unit sphere as the basis point set for all link point clouds:

\vspace{-2mm}
\begin{equation}
    \mathbf{B} = [b_1, ..., b_B]^T, \ ||b_j|| \leq1, \ \forall j.
\end{equation}

Then, the BPS feature can be formulated as the minimum distance between the normalized link point cloud and basis point set to represent the link geometry, which is then encoded to link nodes as illustrated in Sec.~\ref{sec:3.1}.

\vspace{-4mm}
\begin{equation}
    \text{BPS}(P_i^R) = [\text{min}_{j} ||p_{ij}-b_1||, \ldots, \text{min}_{j} ||p_{ij}-b_B||].
\end{equation}

\subsubsection{Graph Denoising Layer} 
\begin{figure}[H]
    \centering
    \includegraphics[width=0.9\linewidth]{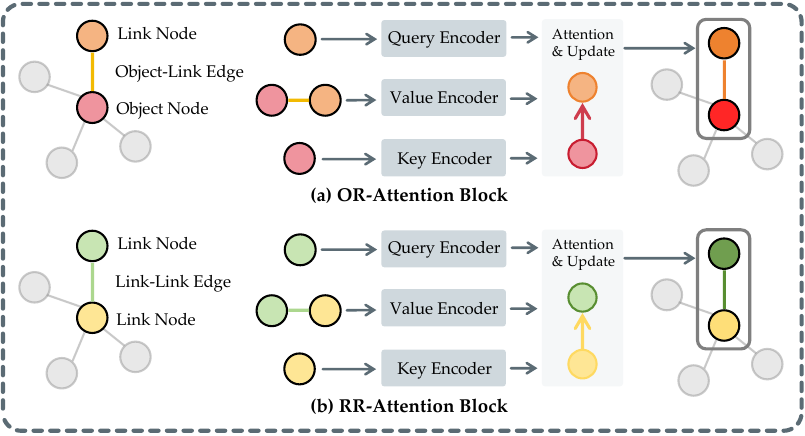}
    \caption{OR and RR attention block.}
    \label{fig:attention}
\end{figure}

To predict noise on link node from the noisy $\mathcal{T(R,O)}$ Graph, we employ a graph denoiser composed of $N=6$ layers, each consisting of one OR-attention and one RR-attention block. Fig.~\ref{fig:attention} illustrates the structure details of both attention blocks, where attention mechanism aggregates information from graph nodes and edges to update their representations.

\vspace{-10pt}

\subsection{Cross-embodiment Zero-shot Generalization}
\label{sec:6.3}

\begin{figure}[H] \centering
   \includegraphics[width=0.98\linewidth]{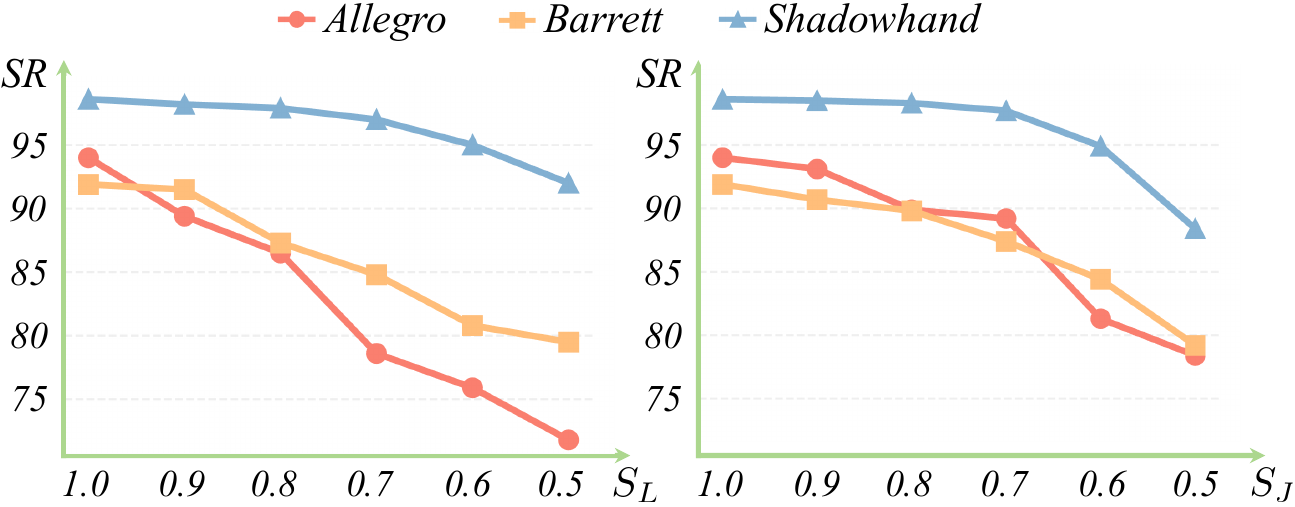}
   \caption{Cross-embodiment zero-shot performance.}
   \label{fig:zero shot}
   \vspace{-20pt}
\end{figure}

Since $\mathcal{T(R,O)}$ Grasp has only been trained on a limited set of dexterous hands, directly performing zero-shot experiments on a completely unseen hand is infeasible. Instead, creating derived embodiments from existing hands provides a controllable way to approximate cross-embodiment zero-shot generalization. Hence, we construct new hand embodiments by modifying the link length and joint limits of Allegro, Barrett, and ShadowHand. To assess the embodiment similarity, we define link alignment $S_L$ and joint overlap $S_J$ as:

\vspace{-10pt}
\begin{equation}
S_L = 1 - \frac{1}{L}\sum\nolimits_{i=1}^L\frac{|l_i'-l_i|}{l_i},
S_J = \frac{1}{L}\sum\nolimits_{i=1}^L \frac{|j_i'\cap j_i|}{|j_i'\cup j_i|}.
\end{equation}

where $l_i$ and $l_i'$ denote the original and modified link lengths, $j_i$ and $j_i'$ are the corresponding joint ranges. We train $\mathcal{T(R,O)}$ Grasp on the original embodiment of Allegro, Barret and Shadowhand, and evaluate it on their derived embodiments. As illustrated in Fig.~\ref{fig:zero shot}, $\mathcal{T(R,O)}$ Grasp achieves over 70\% success rate on test hands with similarity $\geq 0.5$, highlighting the strong zero-shot capability of $\mathcal{T(R,O)}$ across dexterous hand embodiments with comparable geometries. Notably, the current zero-shot performance is constrained by the limited training embodiments. This suggests that, when trained on a large-scale embodiment dataset, $\mathcal{T(R,O)}$ has the potential to scale up to a foundation model for dexterous grasping with strong zero-shot generalization.

\begin{figure*}[!t]
    \centering
    \includegraphics[width=0.93\linewidth]{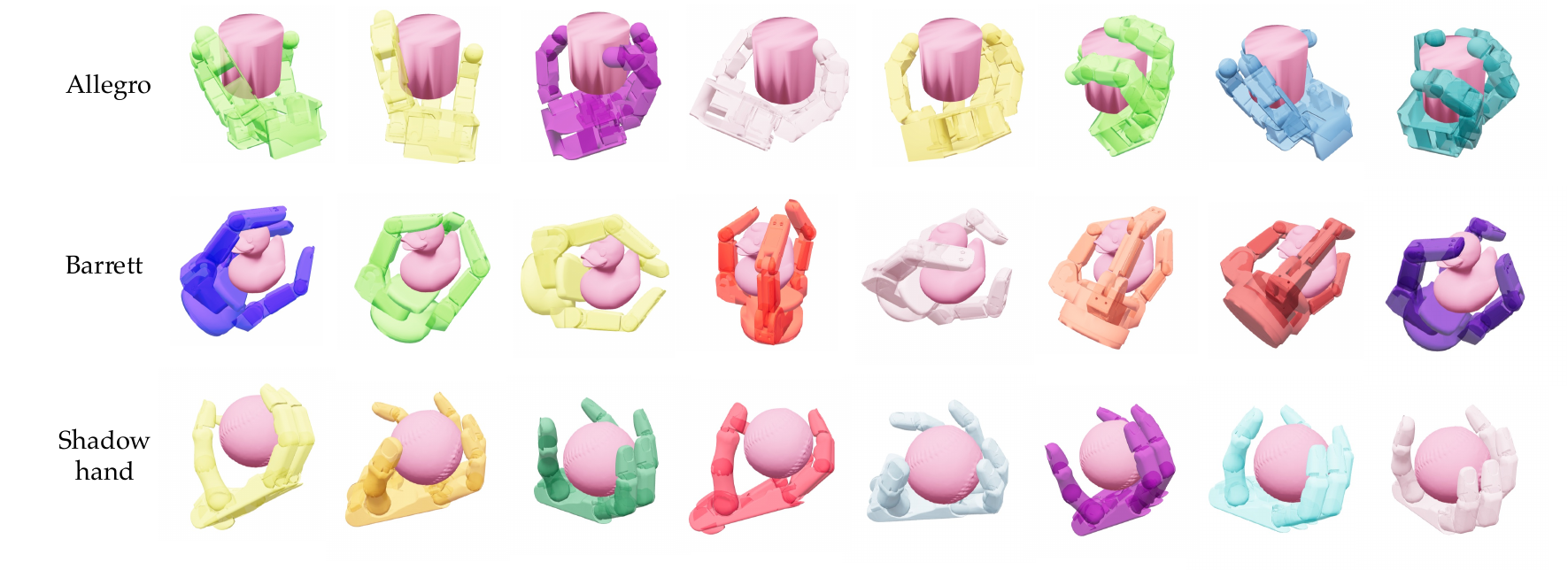}
    \caption{Visualization of unconditioned grasp synthesis.}
    \vspace{-5pt}
    \label{fig:unconditioned suppl}
\end{figure*}

\begin{figure*}[!t]
    \centering
    \includegraphics[width=0.98\linewidth]{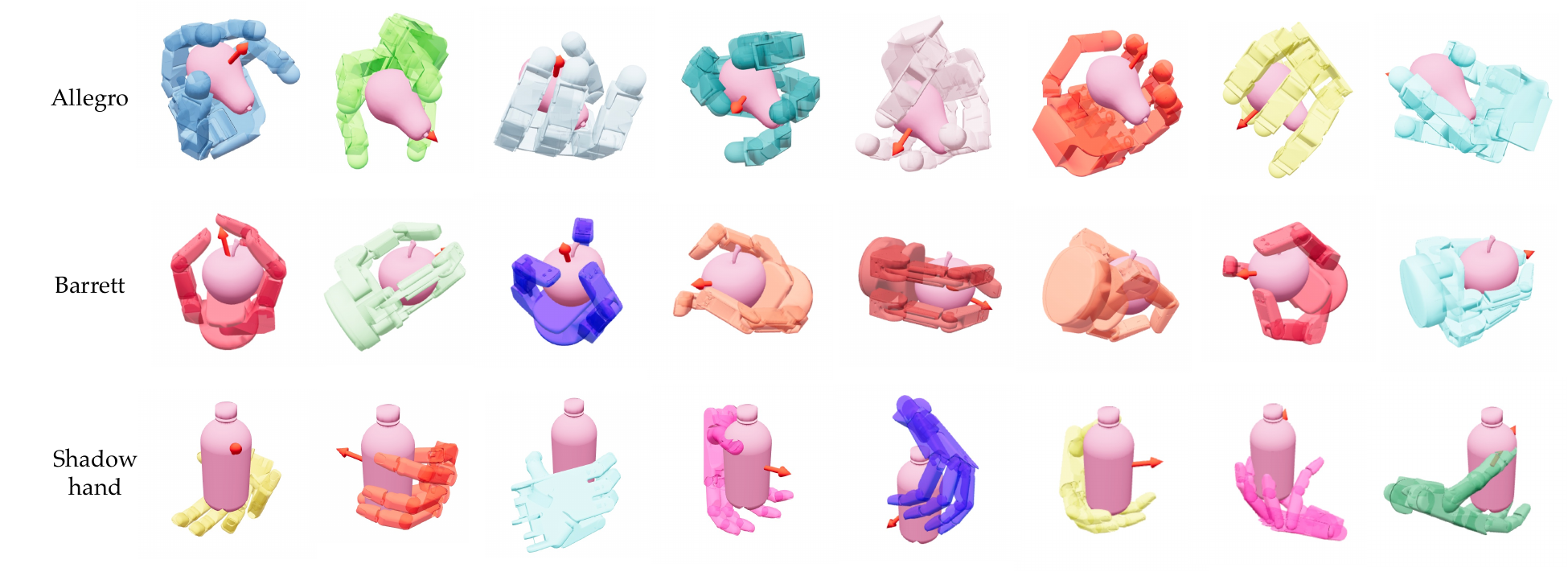}
    \caption{Visualization of conditioned grasp synthesis: red arrow denotes the palm direction of pose guidance.}
    \vspace{-13pt}
    \label{fig:conditioned suppl}
\end{figure*}

\subsection{More Implementation Details}
\label{sec:6.4}

In this section, we provide more comprehensive details on network architecture, training and inference.

\subsubsection{$\mathcal{T(R,O)}$ Graph construction} For object nodes, the pretrained VQ-VAE~\cite{van2017neural} encodes object point cloud into $P=25$ tokens with $64$ dimensions, resulting in object nodes $N^O \in \mathbb{R}^{25 \times (3+1+64)}$. For link nodes, BPS features along with link centers and scales are embeded to 128 dimensions. To allow parallel computing across embodiments with different number of links, link nodes are zero-padded to $N^R \in \mathbb{R}^{25 \times (6+128)}$. Consequently, the object–link edges and link–link edges take the forms $E^{OR} \in \mathbb{R}^{6 \times 25 \times 25}$ and $E^{RR} \in \mathbb{R}^{6 \times 25 \times 24/2}$, respectively.

\subsubsection{Training} During training, we adopt a linear scheduler for the noise variance ranging from $\beta_{\text{min}}=1\times10^{-4}$ to $\beta_{\text{max}}=0.02$ with $T=1000$ diffusion steps in total. $\mathcal{T(R,O)}$ Grasp model is trained for 300 epochs with Adam optimizer. The initial learning rate is set to $1\times10^{-4}$ and decays by a factor of 0.8 every 20 epochs. Position and rotation noise loss weights are set to $\gamma_{p}=\gamma_{r}=1.0$.

\subsubsection{Inference} For both unconditioned and conditioned grasp synthesis, we follow DDIM~\cite{song2020denoising} diffusion strategy to sample $M=20$ steps for inference. To encourage grasp diversity, we set $\lambda=0.2$ to inject random noise during inference. In conditioned grasp synthesis, the strength of the gradient guidance is formulated as:

\vspace{-5pt}
\begin{equation}
s(t)=0.5 \ \text{sin}(\frac{i\pi}{2M}), \ i=1,\ldots, M.
\end{equation}

where $M$ is the total number of inference steps. The schedule $s(t)$ preserves diffusion diversity in the early steps, while encouraging the generated grasp to follow the orientation guidance in the later steps.

\subsection{More Visualization}
\label{sec:6.5}

\begin{figure}[t]
    \centering
    \includegraphics[width=\linewidth]{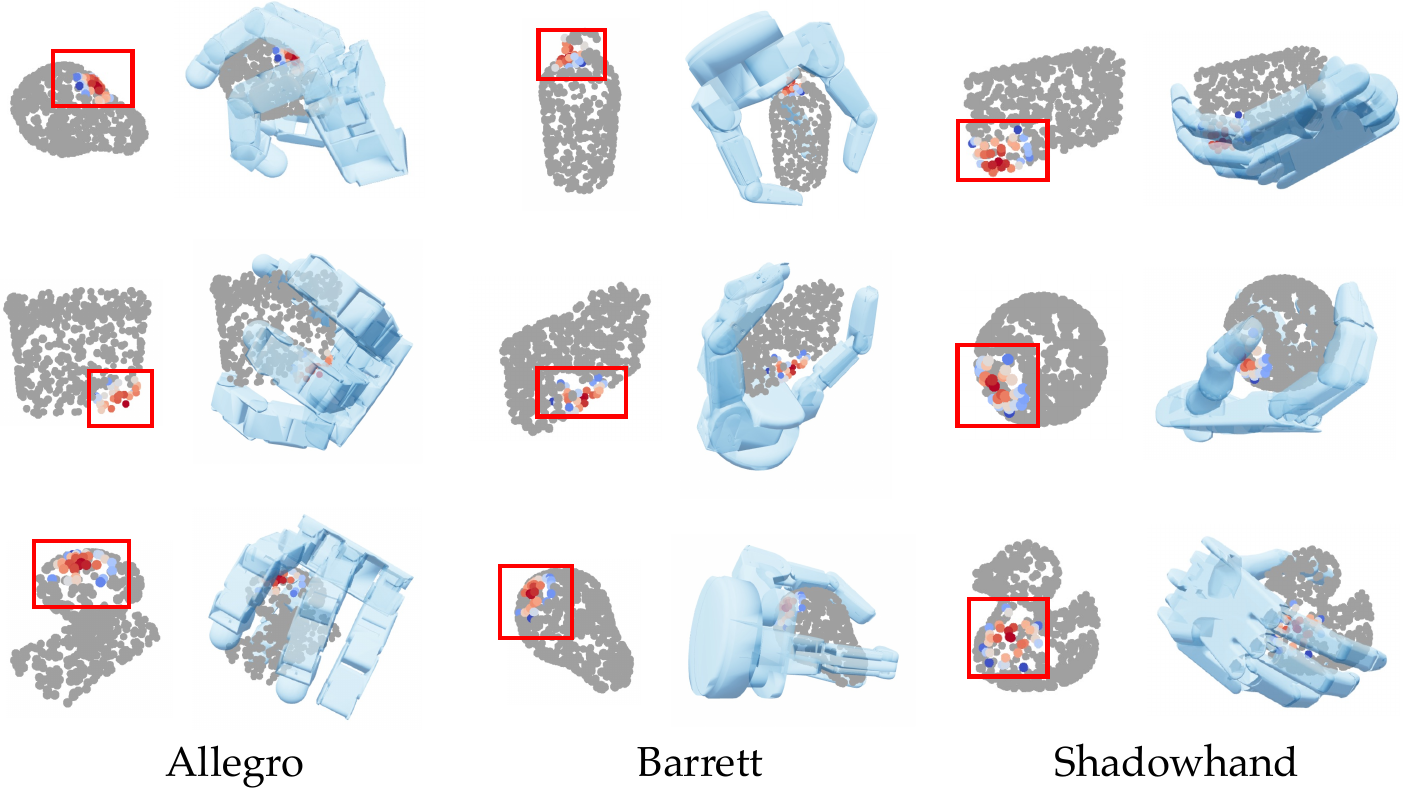}
    \caption{Visualization of conditioned grasp synthesis: red box denotes the contact region guidance.}
    \vspace{-15pt}
    \label{fig:contact suppl}
\end{figure}

We provide more visualization on unconditioned and conditioned grasp synthesis of $\mathcal{T(R,O)}$ Grasp in Fig.~\ref{fig:unconditioned suppl}, Fig.~\ref{fig:conditioned suppl} and Fig.~\ref{fig:contact suppl}, respectively. Our model consistently produces accurate dexterous grasps on novel objects across all embodiments. Moreover, the introduction of pose or contact guidance flexibly controls the grasp synthesis process, while significantly enhancing grasp diversity.

\vspace{-5pt}

\subsection{Discussion and Limitation}
\label{sec:6.6}

$\mathcal{T(R,O)}$ Grasp is an efficient graph diffusion model for cross-embodiment dexterous grasp synthesis. It significantly outperforms all baselines on grasp success rate and inference speed with much lower memory consumption. Leveraging the nature of diffusion models, $\mathcal{T(R,O)}$ Grasp supports both unconditioned and conditioned grasp synthesis, such as conditioning on initial poses or contact regions.

However, $\mathcal{T(R,O)}$ Grasp has certain limitations. As discussed in Sec.~\ref{sec:6.3}, due to the limited set of dexterous hands in training, our model struggles to perform zero-shot grasp synthesis on entirely unseen embodiments. Future work should therefore focus on scaling $\mathcal{T(R,O)}$ Grasp to a broader set of dexterous hands with more diverse and comprehensive geometries. Moreover, $\mathcal{T(R,O)}$ Grasp network does not explicitly account for object–robot collisions or joint limit constraints, which may result in physically infeasible grasp synthesis results. Hence, integrating collision detection and kinematic constraints awareness into the diffusion process can further improve the accuracy of grasp synthesis. Ultimately, we envision $\mathcal{T(R,O)}$ as a foundation representation for robot–object interaction, enabling various downstream applications, such as policy learning and planning.

\end{document}